\documentclass{article}
\usepackage{style/foci}
\usepackage{tgpagella}   % body / title serif (TeX Gyre Pagella)
\usepackage{tgheros}     % sans used by cover style 3
\usepackage[utf8]{inputenc}
\usepackage[T1]{fontenc}
\usepackage{microtype}

\usepackage{amsmath}
\usepackage{amssymb}
\usepackage{amsfonts}
\usepackage{amsthm}
\usepackage{mathtools}
\usepackage{mathrsfs}
\usepackage{physics}
\usepackage{braket}
\usepackage{slashed}
\usepackage{nicefrac}
\usepackage{textcomp}
\usepackage{dsfont}
\usepackage{bbm}
\usepackage{bm}

\usepackage{graphicx}
\usepackage{subcaption}
\usepackage[export]{adjustbox}
\usepackage{float}
\usepackage{booktabs}
\usepackage{dcolumn}
\newcolumntype{d}[1]{D{.}{.}{#1}}
\usepackage{bigstrut, tabularx, multirow, makecell, diagbox}
\usepackage{colortbl}
\usepackage{tabularray}
\UseTblrLibrary{booktabs}
\usepackage{threeparttable}
\usepackage{tablefootnote}
\usepackage{fontawesome5}

\usepackage{placeins}
\usepackage{caption}
\usepackage{footnote}
\usepackage{enumitem}
\usepackage{multicol}
\usepackage{xspace}
\usepackage{titletoc}
\usepackage{titlesec}
\usepackage[bottom]{footmisc}
\usepackage{setspace}

\usepackage{wrapfig}
\usepackage{tikz}
\usepackage{quantikz}
\usepackage{dashbox}
\usepackage{mdframed}
\usepackage{marvosym}
\usepackage{pifont}
\usepackage{CJK}
\usepackage{url}

\usepackage[table,x11names]{xcolor}
\usepackage[most]{tcolorbox}
\tcbuselibrary{breakable}
\usetikzlibrary{decorations.pathreplacing, fit}

\definecolor{primaryblue}{HTML}{0066CC}
\definecolor{accentcyan}{HTML}{00D4AA}
\definecolor{warmorange}{HTML}{FF6B35}
\definecolor{deepgray}{HTML}{2C3E50}
\definecolor{lightgray}{HTML}{F8F9FA}
\definecolor{gradientstart}{HTML}{667eea}
\definecolor{gradientend}{HTML}{764ba2}

\definecolor{citecolor}{HTML}{0071bc}
\definecolor{citeblue}{RGB}{0, 113, 188}
\definecolor{linkcolor}{HTML}{9A4D92}
\definecolor{firebrick}{rgb}{0.698,0.133,0.133}

\definecolor{paleviolet}{HTML}{E1EEFC}
\definecolor{CarolinaUltraLight}{HTML}{E7F4FC}
\definecolor{lightgrey}{RGB}{247, 247, 247}
\definecolor{shadecolor}{HTML}{EFEFEF}
\definecolor{lightyellow}{rgb}{1.0, 0.95, 0.7}
\definecolor{lightblue}{rgb}{0.90, 0.95, 1.0}
\definecolor{light-gray}{gray}{0.95}

\definecolor{darkgrey}{rgb}{0.5, 0.5, 0.5}
\definecolor{darkgreen}{rgb}{0, 0.5, 0}
\definecolor{mydarkblue}{rgb}{0,0.08,0.45}
\definecolor{mydarkblue2}{rgb}{0.133, 0.133, 0.698}
\definecolor{echodrk}{HTML}{0099cc}
\definecolor{mymauve}{rgb}{0.58,0,0.82}
\definecolor{midnightblue}{rgb}{0.1,0.1,0.44}
\definecolor{oxfordblue}{rgb}{0.0,0.13,0.28}
\definecolor{prussianblue}{rgb}{0.0,0.19,0.33}
\definecolor{coolteal}{rgb}{0, 0.45, 0.45}
\definecolor{olive}{rgb}{0.1, 0.3, 0}
\definecolor{mypurple}{rgb}{0.5,0,0.5}
\definecolor{almond}{rgb}{0.94, 0.87, 0.8}

\definecolor{blue_ampEncoding}{HTML}{DAE8FC}
\definecolor{green_encoder}{HTML}{D5E8D4}
\definecolor{purple_decoder}{HTML}{E1D5E7}
\definecolor{yellow_measure}{HTML}{FFF2CC}
\definecolor{gray_block}{HTML}{F5F5F5}
\definecolor{pink_dru}{HTML}{FAD9D5}
\definecolor{orange_v}{HTML}{FAD7AC}

\definecolor{colorA}{rgb}{1,0,0}
\definecolor{colorB}{rgb}{0,0.3,1}
\definecolor{colorC}{rgb}{0.9,0.8,0.2}
\definecolor{colorD}{rgb}{0,0.65,0}
\definecolor{lesslightgray}{rgb}{0.5,0.5,0.5}
\definecolor{fundamental}{RGB}{55, 110, 111}
\definecolor{Gred}{RGB}{219, 50, 54}
\definecolor{ToCgreen}{RGB}{0, 128, 0}
\definecolor{Sepia}{RGB}{112, 66, 20}
\definecolor{Dblue}{rgb}{0,0.08,0.45}
\definecolor{Blue}{rgb}{0, 0, 0.8}
\definecolor{blue}{rgb}{0,0,1}
\definecolor{UNCblue!10}{rgb}{0.84,0.91,0.98}
\definecolor{RowAlt}{rgb}{0.98,0.98,0.99}

\definecolor{CarolinaBlue}{HTML}{7BAFD4}        % Official UNC Carolina Blue
\definecolor{CarolinaLightBlue}{HTML}{B3D4E5}   % Lighter version
\definecolor{CarolinaUltraLight}{HTML}{E8F4F8}  % Very light for background
\definecolor{CarolinaText}{HTML}{1C2B33}        % Dark text color

\usepackage[pagebackref=true,breaklinks=true,colorlinks,hyperfootnotes=false]{hyperref}
\hypersetup{
  colorlinks,
  citecolor=citeblue,
  linkcolor=firebrick,
  urlcolor=firebrick
}
\usepackage[nameinlink,capitalize,noabbrev]{cleveref}

\titlespacing\section{0pt}{4pt plus 4pt minus 2pt}{-2pt plus 2pt minus 2pt}
\titlespacing\subsection{0pt}{2pt plus 4pt minus 2pt}{-2pt plus 2pt minus 2pt}
\titlespacing\subsubsection{0pt}{2pt plus 4pt minus 2pt}{-2pt plus 2pt minus 2pt}

\makeatletter
\def\th@remark{%
  \thm@headfont{\bfseries}%
  \normalfont % body font
  \thm@preskip\topsep \divide\thm@preskip\tw@
  \thm@postskip\thm@preskip
}
\makeatother

\theoremstyle{definition}

\tcbset{focithm/.style={
  enhanced, breakable, frame hidden,
  colback=black!5,
  width=\linewidth,
  enlarge left by=0pt, enlarge right by=0pt,
  boxsep=6pt, boxrule=0pt,
  left=4pt, right=4pt, top=1pt, bottom=1pt,
  arc=8pt,
  before skip=\topsep, after skip=\topsep
}}

\tcolorboxenvironment{theorem}{focithm}

\tcolorboxenvironment{lemma}{focithm}

\tcolorboxenvironment{corollary}{focithm}

\tcolorboxenvironment{proposition}{focithm}

\tcolorboxenvironment{definition}{focithm}

\tcolorboxenvironment{assumption}{focithm}

\tcolorboxenvironment{claim}{focithm}

\tcolorboxenvironment{problem}{focithm}

\tcolorboxenvironment{question}{focithm}

\newtcolorbox{titleblock}{
  enhanced,
  frame hidden,
  colback=CarolinaUltraLight,
  colframe=CarolinaUltraLight,
  boxrule=0pt,
  arc=10pt,
  left=14pt,
  right=14pt,
  top=14pt,
  bottom=14pt,
  width=\linewidth,
  before skip=12pt plus 4pt,
  after skip=12pt plus 4pt,
  grow to left by=1.5pt,
  grow to right by=1.5pt,
  before upper={
    \setlength{\parindent}{0cm}
    \setlength{\parskip}{0.5cm}
  }
}

\crefname{theorem}{Theorem}{Theorems}
\crefname{proposition}{Proposition}{Propositions}
\crefname{lemma}{Lemma}{Lemmas}
\crefname{corollary}{Corollary}{Corollaries}
\crefname{definition}{Definition}{Definitions}
\crefname{assumption}{Assumption}{Assumptions}
\crefname{remark}{Remark}{Remarks}
\crefname{problem}{Problem}{Problems}
\crefname{property}{Property}{property}
\crefname{question}{Question}{Questions}

\numberwithin{equation}{section}
\numberwithin{theorem}{section}
\numberwithin{proposition}{section}
\numberwithin{definition}{section}
\numberwithin{lemma}{section}
\numberwithin{assumption}{section}
\numberwithin{remark}{section}

\newcommand{\best}[1]{\mathbf{#1}}

\def\1{\bm{1}}

\makeatletter
\let\save@mathaccent\mathaccent
\newcommand*\if@single[3]{%
    \setbox0\hbox{${\mathaccent"0362{#1}}^H$}%
    \setbox2\hbox{${\mathaccent"0362{\kern0pt#1}}^H$}%
    \ifdim\ht0=\ht2 #3\else #2\fi
}
\newcommand*\rel@kern[1]{\kern#1\dimexpr\macc@kerna}
\newcommand*\widebar[1]{\@ifnextchar^{{\wide@bar{#1}{0}}}{\wide@bar{#1}{1}}}
\newcommand*\wide@bar[2]{\if@single{#1}{\wide@bar@{#1}{#2}{1}}{\wide@bar@{#1}{#2}{2}}}
\newcommand*\wide@bar@[3]{%
    \begingroup
    \def\mathaccent##1##2{%
        \let\mathaccent\save@mathaccent
        \if#32 \let\macc@nucleus\first@char \fi
        \setbox\z@\hbox{$\macc@style{\macc@nucleus}_{}$}%
        \setbox\tw@\hbox{$\macc@style{\macc@nucleus}{}_{}$}%
        \dimen@\wd\tw@
        \advance\dimen@-\wd\z@
        \divide\dimen@ 3
        \@tempdima\wd\tw@
        \advance\@tempdima-\scriptspace
        \divide\@tempdima 10
        \advance\dimen@-\@tempdima
        \ifdim\dimen@>\z@ \dimen@0pt\fi
        \rel@kern{0.6}\kern-\dimen@
        \if#31
        \overline{\rel@kern{-0.6}\kern\dimen@\macc@nucleus\rel@kern{0.4}\kern\dimen@}%
        \advance\dimen@0.4\dimexpr\macc@kerna
        \let\final@kern#2%
        \ifdim\dimen@<\z@ \let\final@kern1\fi
        \if\final@kern1 \kern-\dimen@\fi
        \else
        \overline{\rel@kern{-0.6}\kern\dimen@#1}%
        \fi
    }%
    \macc@depth\@ne
    \let\math@bgroup\@empty \let\math@egroup\macc@set@skewchar
    \mathsurround\z@ \frozen@everymath{\mathgroup\macc@group\relax}%
    \macc@set@skewchar\relax
    \let\mathaccentV\macc@nested@a
    \if#31
    \macc@nested@a\relax111{#1}%
    \else
    \def\gobble@till@marker##1\endmarker{}%
    \futurelet\first@char\gobble@till@marker#1\endmarker
    \ifcat\noexpand\first@char A\else
    \def\first@char{}%
    \fi
    \macc@nested@a\relax111{\first@char}%
    \fi
    \endgroup
    }
\makeatother

\DeclareMathAlphabet{\mathsfit}{\encodingdefault}{\sfdefault}{m}{sl}
\SetMathAlphabet{\mathsfit}{bold}{\encodingdefault}{\sfdefault}{bx}{n}

\renewcommand{\arraystretch}{1.15}
\makeatletter

\def\foci@authors{}
\def\foci@affil{}
\def\foci@email{}
\def\foci@project{}
\def\foci@date{\today}
\def\foci@authorsep{}

\newcommand{\fociauthor}[2]{%
  \g@addto@macro\foci@authors{\foci@authorsep\mbox{#1\textsuperscript{#2}}\gdef\foci@authorsep{,\ }}}
\newcommand{\fociaffil}[1]{\gdef\foci@affil{#1}}
\newcommand{\fociemail}[1]{\gdef\foci@email{#1}}
\newcommand{\fociproject}[1]{\gdef\foci@project{#1}}
\newcommand{\focidate}[1]{\gdef\foci@date{#1}}
\newcommand{\fociauthorlist}{\gdef\foci@authorsep{}\foci@authors}
\newcommand{\focicorr}{\textrm{\Letter}}
\newcommand{\focimarkeq}{*}

\def\foci@blfootnote{\gdef\@thefnmark{}\@footnotetext}
\newcommand{\focifootnotes}{%
  \foci@blfootnote{\textsuperscript{*}Equal contribution.\quad
    \textsuperscript{\Letter}Corresponding author: \texttt{\foci@email}}}

\newlength{\foci@mw}
\newcommand{\focidrawmark}[3]{%
  \setlength{\foci@mw}{#2}%
  \begin{scope}[shift={#3}, x={\foci@mw/564.03}, y={-\foci@mw/564.03}]
    \foreach \cx/\cy in {282.02/66.03, 282.02/222.08, 215.77/348.8, 348.73/348.8,
                         414.88/474.93, 149.16/474.93, 67.06/222.08, 496.99/222.08}
      \fill[#1] (\cx,\cy) circle[radius=42.04];
  \end{scope}}
\newcommand{\focimark}[2][black]{%
  \setlength{\foci@mw}{#2}%
  \begin{tikzpicture}[x={\foci@mw/564.03}, y={-\foci@mw/564.03}]
    \useasboundingbox (0,0) rectangle (564.03,540.96);
    \focidrawmark{#1}{#2}{(0,0)}
  \end{tikzpicture}}

\fancypagestyle{focititlepage}{\fancyhf{}\renewcommand{\headrulewidth}{0pt}}

\newcommand{\foci@headerline}{%
  \noindent
  \begin{minipage}[c]{0.5\linewidth}
    \includegraphics[height=0.58cm]{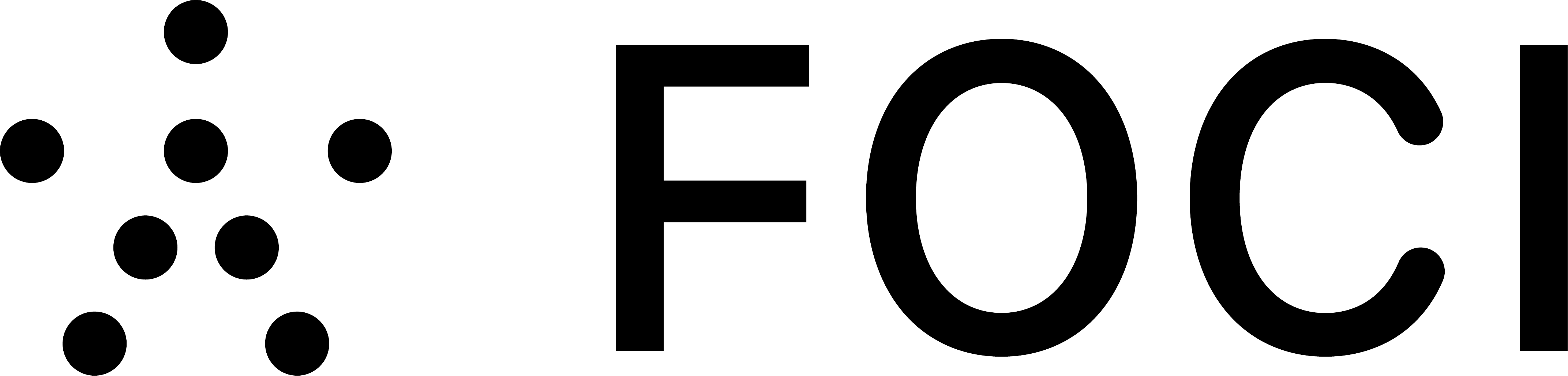}
  \end{minipage}%
  \begin{minipage}[c]{0.5\linewidth}
    \raggedleft\small\itshape\color{black!55}\foci@date
  \end{minipage}\par}

\newcommand{\foci@contactline}{%
  {\footnotesize\color{black!65}%
   \textbf{Email:}~\texttt{\foci@email}\qquad
   \textbf{Project:}~\url{\foci@project}\par}}

\newcommand{\foci@hairline}{\noindent\textcolor{black}{\rule{\linewidth}{0.5pt}}\par}

\newcommand{\focicoverCard}{%
  \thispagestyle{focititlepage}%
  \foci@headerline
  \vskip -0.5em
  \foci@hairline
  \vskip 0.9em
  \begin{tcolorbox}[enhanced, frame hidden, boxrule=0pt, arc=12pt,
      colback=black!4, width=\linewidth,
      left=18pt, right=18pt, top=18pt, bottom=16pt,
      grow to left by=2pt, grow to right by=2pt]
    \noindent
    \begin{minipage}[t]{\linewidth}
      {\setstretch{1.15}\LARGE\bfseries\raggedright Aligning Multimodal Patient Evidence with Biomedical Knowledge Graphs for Clinical LLMs
\unskip\par}
      \vskip 0.55em
      {\normalsize\raggedright\fociauthorlist\par}
      \vskip 0.2em
      {\small\color{black!70}\raggedright\foci@affil\par}
    \end{minipage}\par
    \vskip 0.9em
    {\small\setstretch{1.08}Clinical questions often depend on linking a patient's multimodal evidence to external biomedical knowledge, yet existing predictive systems rarely represent such links explicitly, so they can neither be traced to their evidence sources nor removed to measure their contributions.
We present \graphname{} (\textbf{M}ulti\textbf{m}odal \textbf{K}nowledge \textbf{G}raph), which represents heterogeneous, multimodal patient observations and biomedical concepts as separate layers in one typed graph, joined by explicit alignment edges.
First, modality-specific harmonizers convert EHR text, imaging, genomic, and biospecimen data into
typed observations mapped to UMLS concepts, which a route-prioritized aligner links to a
biomedical knowledge graph.
Query-conditioned retrieval then selects a compact subgraph for downstream use by a large language model or a graph neural network.
We build MM-KGs for MIMIC-IV and ADNI, and
evaluate them with a $2\times 2$ design that separates patient evidence, biomedical
knowledge, and their interaction.
On questions that require both sources, neither source alone performs far above chance,
whereas their combination yields a drug-controlled AUROC interaction of $+0.194$ on MIMIC
and $+0.299$ on ADNI.
On held-out five-candidate ranking, \graphname{} outperforms MindMap by $+0.131$ Hits@1 and
leads an adapted GraphCare on the items that require consulting the patient, and deleting
the single answer-bearing relation from the retrieved packet returns Hits@1 to the no-knowledge baseline.
Finally, query-conditioned retrieval reaches $0.731$ AUROC with $6.8\times$ less context than the
strongest generic policy, whereas static knowledge graph context gives no consistent gain on
ordinary outcome prediction.
Knowledge graphs thus benefit clinical LLMs not as background context but as explicit links between multimodal patient evidence and the relation a question requires, and \graphname{} makes these links retrievable, traceable, and testable.
\unskip\par}
    \vskip 0.9em
    \foci@contactline
  \end{tcolorbox}
  \vskip 0.4em}

\newcommand{\focicover}{\focicoverCard}

\makeatother

\usepackage{fvextra}
\DefineVerbatimEnvironment{PromptVerbatim}{Verbatim}{
  breaklines=true,
  breakanywhere=true,
  fontsize=\footnotesize,
  breaksymbolleft={}
}
\usetikzlibrary{positioning,fit}
\graphicspath{{figures/}}
\definecolor{cellhi}{RGB}{226,243,239}
\definecolor{inkboth}{RGB}{14,138,115}
\definecolor{inkmm}{RGB}{46,111,158}
\definecolor{inkkg}{RGB}{140,91,166}

\newcommand{\graphname}{MM-KG}
\renewcommand{\best}[1]{\textbf{#1}} % style/macros.tex defines a math-mode \best

\providecommand{\cyes}{\ding{51}}
\providecommand{\cno}{\ding{55}}
\providecommand{\cpart}{$\sim$}
\hypersetup{
  pdftitle={Aligning Multimodal Patient Evidence with Biomedical Knowledge Graphs for Clinical LLMs},
  pdfauthor={Jiawen Du, Arshan Ali Khan, Chenhao Zhang, Zachary Plotkin, Li Shen, Qi Long, Yun Li, Can Chen, Tianlong Chen, Nicholas Konz}
}

\begin{document}

% Cover date: defaults to today's date. To pin a specific date instead,
% uncomment the next line and edit it.
% \focidate{June 30, 1905}
% One author per \fociauthor; use \focimarkeq for equal contribution,
% \focicorr for corresponding author.
\fociauthor{Jiawen Du}{1}
\fociauthor{Arshan Ali Khan}{2,\focimarkeq}
\fociauthor{Chenhao Zhang}{\focimarkeq}
\fociauthor{Zachary Plotkin}{3}
\fociauthor{Li Shen}{4}
\fociauthor{Qi Long}{4}
\fociauthor{Yun Li}{1}
\fociauthor{Can Chen}{1}
\fociauthor{Tianlong Chen}{1,\focicorr}
\fociauthor{Nicholas Konz}{1,\focicorr}

\fociaffil{%
  \mbox{\textsuperscript{1}University of North Carolina at Chapel Hill (unc.edu)}\quad
  \mbox{\textsuperscript{2}Siddaganga Institute of Technology (sit.ac.in)}\quad
  \mbox{\textsuperscript{3}Foci Labs (fo.ci)}\quad
  \mbox{\textsuperscript{4}University of Pennsylvania (upenn.edu)}}
\fociemail{\{tianlong, nickk124\}@cs.unc.edu}

% Two corresponding authors, no project page, and the ADNI byline footnote
% (ADNI Data Use Agreement) in place of the template's single-author defaults.
\makeatletter
\renewcommand{\focifootnotes}{%
  \foci@blfootnote{\textsuperscript{*}Equal contribution.\quad
    \textsuperscript{\Letter}Corresponding authors: \texttt{\foci@email}}%
}
\renewcommand{\foci@contactline}{%
  {\footnotesize\color{black!65}\textbf{Email:}~\texttt{\foci@email}\par}}
\makeatother

% running head / foot for pages 2+
\pagestyle{fancy}
\fancyhf{}
\renewcommand{\headrulewidth}{0.4pt}
\chead{\small\scshape Aligning Multimodal Patient Evidence with Biomedical Knowledge Graphs for Clinical LLMs
\unskip}
\cfoot{\small\thepage}

\focicover
\focifootnotes

\section{Introduction}

Reasoning about an individual clinical case requires connecting heterogeneous patient
observations to general biomedical knowledge.
Patient records describe \textit{observations}, e.g., laboratory measurements, images, reports,
medications, longitudinal events \citep{johnson2023mimiciv,johnson2019mimiccxrjpg};
whereas biomedical knowledge bases describe general relations among diseases, drugs, genes,
and phenotypes \citep{chandak2023primekg,bodenreider2004umls,stear2024petagraph}.
Neither is sufficient alone when a question depends on a link between the two, and such
questions are common: across 28{,}582 MIMIC-IV ICU stays, 29.9\% exhibit a laboratory-evident
organ state that never appears in the coded diagnoses while the patient receives a medication
that PrimeKG lists as contraindicated for that state (App.~\ref{app:cohort}).
Answering such questions correctly requires reasoning about the current measurement \emph{and} the drug-specific
relation; either source in isolation leaves the question undecidable.

Existing works model each of these sides, but rarely their connection.
Multimodal clinical models fuse EHR data, text, and images into predictive representations, but
discard the links from individual observations to biomedical concepts
\citep{hayat2022medfuse,yao2024drfuse,yao2024ddlcxr}.
Knowledge-graph methods supply external facts or build patient-specific graphs for prediction and language-model reasoning \citep{jiang2024graphcare,jiang2025kare,wen2024mindmap,wu2024medgraphrag}.
Their patient side, however, is typically derived from structured codes or clinical text alone;
imaging, genomic, and biospecimen evidence is either excluded or fused outside the graph, so no single representation carries all modalities through to the biomedical knowledge backbone.
Where multiple modalities are represented graphically, they are generally not linked to standardized biomedical concepts \citep{wu2024muse}.
Consequently, no explicit path runs from a source observation, through its normalized concept,
to the biomedical relation used to interpret it, and the connection cannot be retrieved or ablated.

\begin{figure}[t]
    \centering
    \includegraphics[width=0.65\textwidth]{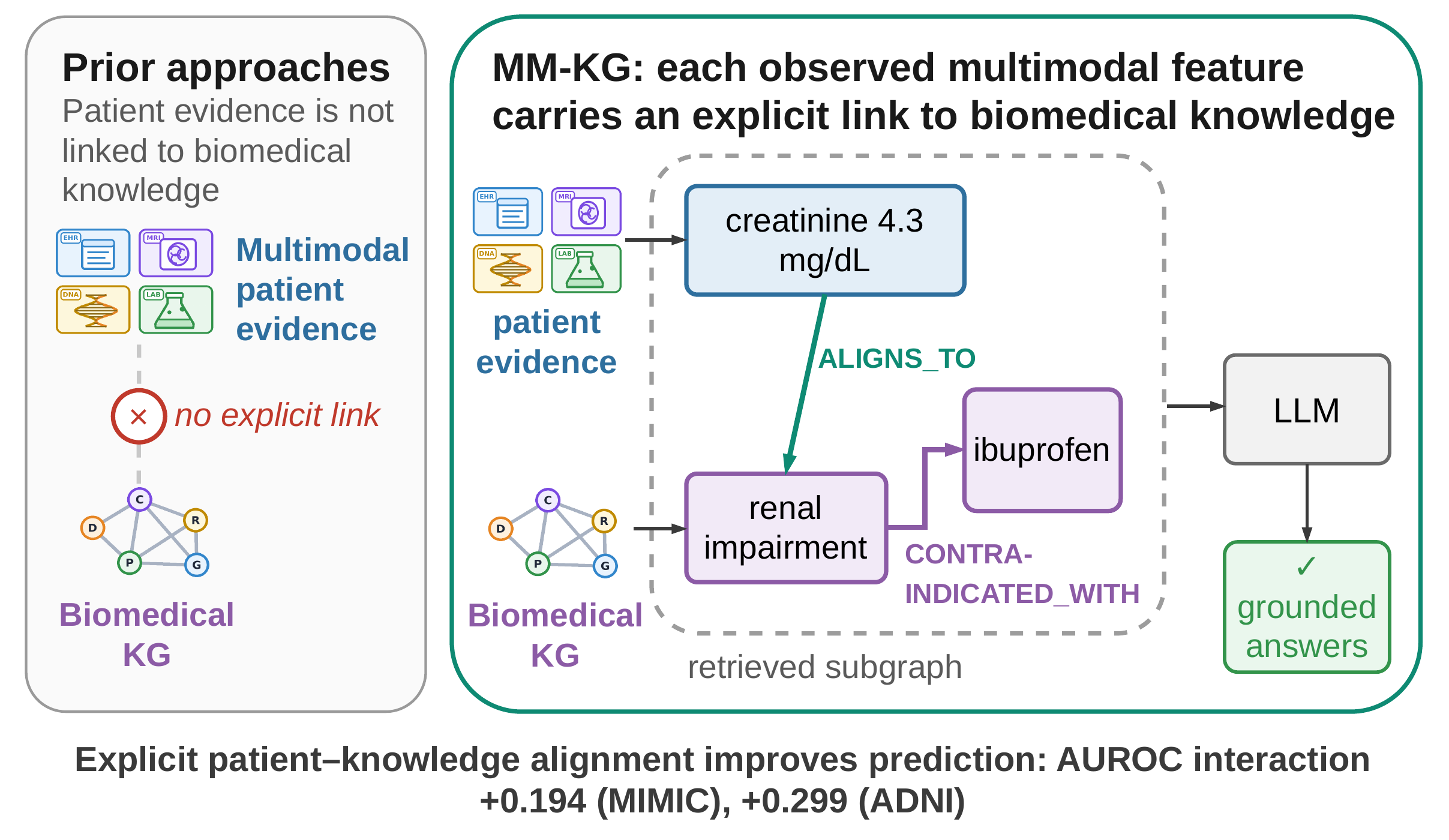}
    \caption{\textbf{Motivation of \graphname{}.}
    %The displayed values are the controlled AUROC interactions for MIMIC and held-out ADNI validation.
    }
    \label{fig:teaser}
\end{figure}

We introduce \graphname{} (\textbf{M}ulti\textbf{m}odal \textbf{K}nowledge \textbf{G}raph), which represents multimodal patient evidence and a biomedical knowledge backbone in a single typed graph.
Each modality passes through its own harmonizer, converting raw input into typed
observation nodes rather than a fused vector.
A vision-language model turns chest radiographs into calibrated textual findings;
brain MRIs are segmented into regional abnormality findings under a normative
model; and codes, measurements, and report text are resolved by vocabulary
lookup, value-aware rules, and entity linking.
Provenance edges retain each observation's source and timestamp, and a route-prioritized aligner links the resulting concepts to PrimeKG \citep{chandak2023primekg}, preferring exact identifier routes over synonym and parent hops (\S\ref{sec:alignment}).
Because instance and ontology nodes keep separate identities, each layer can be traced or
removed independently.
Query-conditioned retrieval then assembles a compact subgraph per question, joining the relevant patient neighborhood to candidate-conditioned domain relations under deterministic rules with no access to the label.
The result can then be serialized for a downstream LLM or used as input to a GNN, exposing a path from source observations, to aligned concepts, to the relation supporting an answer (Fig.~\ref{fig:teaser}).
This explicit structure makes the patient evidence, biomedical knowledge, and their alignment separately traceable and testable.

We build two independently constructed graphs: 51{,}844 MIMIC-IV patients with chest radiographs and reports, and 1{,}000 ADNI participants spanning longitudinal imaging, history, cognition, genetics, and biospecimens \citep{veitch2022adni}.
Because instance and ontology layers remain separately addressable, a $2{\times}2$ design can isolate patient evidence, KG relations, and their interaction (\S\ref{sec:factorial}), a decomposition which fused representations do not support.
Our primary evaluation targets questions in which neither source is sufficient alone.
On 3{,}200 controlled MIMIC questions that link a condition documented in laboratory values or radiology reports to a drug relation, each source alone leaves performance near chance, whereas together they yield a drug-controlled AUROC interaction of $+0.194$ ($p{<}0.001$), reproduced on the 70-item ADNI test set at $+0.299$ across a different disease domain and evidence channel.
On held-out five-candidate ranking, \graphname{} exceeds MindMap by $+0.131$ Hits@1 and leads an adapted GraphCare on the items a candidate-only prior cannot resolve (0.582 against 0.375), while deleting the answer-bearing relation, replacing or shuffling packets reduces Hits@1 from 0.575 to 0.185--0.299 against 0.200 chance.
The effect therefore requires the specific correspondence that alignment makes explicit, which query-conditioned retrieval supplies at AUROC 0.731 with 1{,}206 input tokens, against 8{,}157 for the strongest generic policy.
Where one source suffices, each behaves as expected: patient measurements drive mortality prediction for both consumers while static PrimeKG adds nothing measurable (App.~\ref{app:noise}), and the same relations improve biomedical fact-checking by $+0.063$ to $+0.420$ (App. Table~\ref{tab:audit}).

\textbf{Our contributions are summarized as follows.}
\textbf{(1)}~A typed instance--knowledge graph construction framework in which multimodal observations, their provenance, and their alignment to a biomedical backbone remain separately addressable, so that each layer can be traced or ablated rather than fused away.
\textbf{(2)}~Query-conditioned retrieval that joins patient evidence to candidate-conditioned relations, recovering the answer path for 99.7\% of positive items and reaching AUROC 0.731 with $6.8\times$ less context than the strongest vector, full-context, or typed one-hop policy.
\textbf{(3)}~A factorial evaluation showing that patient evidence and KG relations combine superadditively on questions that require both, yielding interactions of $+0.194$ (MIMIC) and $+0.299$ (ADNI) AUROC, and that removing or corrupting the answer-bearing alignment returns held-out ranking to the no-knowledge level or below, whereas static KG context alone does not improve outcome prediction.
\graphname{} construction, retrieval, benchmark, and evaluation code will be released upon acceptance.\footnote{The KGs themselves are reproducible from credentialed, non-redistributable MIMIC-IV and ADNI data.}

\section{Related work}
\label{sec:related}

\textbf{Multimodal clinical representation.}
Clinical representation learning has progressed from longitudinal EHR models such as RETAIN \citep{choi2016retain} and GRAM \citep{choi2017gram} to multimodal architectures that combine structured records, text, and medical images.
MedFuse \citep{hayat2022medfuse}, DrFuse \citep{yao2024drfuse}, and DDL-CXR study fusion \citep{yao2024ddlcxr} under missing or asynchronous modalities.
MedGTX \citep{park2022medgtx} jointly encodes structured EHR graphs and clinical text, while MUSE \citep{wu2024muse} represents patient--modality relationships with a bipartite graph and evaluates on both MIMIC and ADNI. 
MedMod \citep{elsharief2025medmod}  provides a benchmark for multimodal prediction using MIMIC-IV and MIMIC-CXR.
These methods are important prediction baselines, but they do not explicitly preserve the paths from source observations through normalized biomedical concepts to external biomedical relations.

\textbf{Patient representations and biomedical KGs.}
Multimodal KGs associate symbolic entities with heterogeneous attributes or evidence \citep{chen2024mmkgsurvey}.
Biomedical resources such as PrimeKG and PetaGraph \citep{chandak2023primekg,stear2024petagraph} provide large concept-level backbones but do not by themselves represent a patient's multimodal clinical history.
Several methods enrich patient or code representations with biomedical knowledge.
MMUGL \citep{burger2023mmugl} maps structured codes and concepts extracted from clinical text into a shared UMLS-derived representation;  ClinVec and MedTok combine standardized codes with textual or relational knowledge \citep{johnson2024clinvec,su2025medtok};and GraphCare constructs patient-specific graphs using external biomedical resources \citep{jiang2024graphcare}. 
KG4Diagnosis proposes a hierarchical multi-agent framework that constructs a graph from unstructured medical text for diagnostic reasoning \citep{zuo2025kg4diagnosis}.
These studies show the value of combining patient records with biomedical knowledge; \graphname{} builds on this further by retaining source-level multimodal observations, recording how each was derived, and linking the resulting concepts explicitly to a shared biomedical backbone.

\textbf{Knowledge retrieval for clinical prediction.}
Retrieval-augmented generation (RAG)  \citep{lewis2020rag} retrieves external evidence before producing an answer.
In clinical prediction, RAM-EHR retrieves relevant information from external knowledge sources \citep{xu2024ramehr}, while KARE organizes knowledge into communities and retrieves community-level evidence for healthcare prediction \citep{jiang2025kare}.
RAD retrieves disease-centered guideline knowledge and uses it to align multimodal features for clinical diagnosis \citep{li2025rad}.
KARE and RAD are especially relevant comparators, but they use retrieved knowledge primarily to construct summaries or guide latent representations. 
In contrast, \graphname{} retrieves an explicit typed subgraph spanning both patient evidence and biomedical relations, which can be consumed by an LLM or a heterogeneous GNN.

\textbf{KG retrieval and verification for medical LLMs.}
Several methods use biomedical graphs to support medical language models.
MindMap \citep{wen2024mindmap} retrieves graph evidence for diagnostic reasoning, and DR.KNOWS \citep{gao2025drknows} incorporates UMLS knowledge into diagnosis prediction.
DOSSIER checks claims against structured patient records and biomedical knowledge graphs while limiting the exposure of private records to the LLM \citep{zhang2024dossier}. 
FAITH instead evaluates generated medical claims by decomposing responses into atomic statements and scoring their support using paths in a biomedical graph \citep{zhou2026faith}.
MedGraphRAG \citep{wu2024medgraphrag} is the closest prior system: it builds a hierarchical graph over MIMIC-IV, anchors entities to standardized vocabularies, retains source attribution, and conditions retrieval on the query.
It operates over document chunks, so attribution is recorded per passage rather than per observation, and its patient side carries no laboratory values, imaging, or genomics.
\graphname{} records provenance for each typed observation across modalities and makes each observation--concept link a separate, removable edge.
Our experiments then address a complementary question: whether separately controlled patient evidence, external biomedical relations, and their explicit alignment improve prediction and grounding before an answer is generated.
Table~\ref{tab:capability} summarizes these distinctions across representative multimodal, knowledge-enhanced, and graph-retrieval methods.
\begin{table}[t]
\centering
\caption{\textbf{Capabilities of related methods compared to \graphname{}.} ``\cpart'': partial. ``Link'': patient observations linked to a standard vocabulary; ``Prov'': per-observation provenance; ``Q-Ret'': question-conditioned retrieval.}
\label{tab:capability}
\fontsize{9.5pt}{9.5pt}\selectfont
\setlength{\tabcolsep}{4.5pt}
\begin{tabular}{l cccccccccc}
\toprule
& \multicolumn{5}{c}{Patient evidence in the graph} & \multicolumn{3}{c}{Representation} & \multicolumn{2}{c}{Consumer} \\
\cmidrule(lr){2-6}\cmidrule(lr){7-9}\cmidrule(lr){10-11}
Method & Codes & Labs & Imaging & Text & Genetic & Link & Prov & Q-Ret & LLM & GNN \\
\midrule
\multicolumn{11}{l}{\emph{Multimodal clinical representation}}\\
MedFuse~\citep{hayat2022medfuse} & \cno   & \cyes & \cyes & \cno   & \cno & \cno   & \cno   & \cno   & \cno  & \cno  \\
MUSE~\citep{wu2024muse} & \cyes  & \cyes & \cyes & \cyes  & \cno & \cno   & \cno   & \cno   & \cno  & \cyes \\
MedGTX~\citep{park2022medgtx} & \cyes  & \cno  & \cno  & \cyes  & \cno & \cno   & \cno   & \cpart & \cno  & \cyes \\
\addlinespace[2pt]
\multicolumn{11}{l}{\emph{Knowledge-enhanced patient representation}}\\
MMUGL~\citep{burger2023mmugl} & \cyes  & \cno  & \cno  & \cyes  & \cno & \cyes  & \cno   & \cno   & \cno  & \cyes \\
GraphCare~\citep{jiang2024graphcare} & \cyes  & \cno  & \cno  & \cno   & \cno & \cyes  & \cpart & \cpart & \cno  & \cyes \\
\addlinespace[2pt]
\multicolumn{11}{l}{\emph{Graph retrieval for clinical prediction and medical LLMs}}\\
KARE~\citep{jiang2025kare} & \cyes  & \cno  & \cno  & \cno   & \cno & \cyes  & \cno   & \cyes  & \cyes & \cno  \\
MindMap~\citep{wen2024mindmap} & \cno   & \cno  & \cno  & \cyes  & \cno & \cpart & \cpart & \cyes  & \cyes & \cno  \\
MedGraphRAG~\citep{wu2024medgraphrag} & \cpart & \cno  & \cno  & \cyes  & \cno & \cyes  & \cyes  & \cyes  & \cyes & \cno  \\
\midrule
\textbf{MM-KG (ours)} & \cyes & \cyes & \cyes & \cyes & \cyes & \cyes & \cyes & \cyes & \cyes & \cyes \\
\bottomrule
\end{tabular}
\end{table}
% \textbf{Position of this work.}
% \graphname{} combines four properties that are usually studied separately: multimodal patient harmonization, source-preserving instance graphs, explicit alignment to a biomedical backbone, and query-conditioned retrieval. 
% We evaluate these components with both LLM and GNN consumers and use controlled ablations to distinguish the contribution of patient evidence, external relations, retrieval, and alignment.
% The closest comparisons are therefore multimodal clinical prediction methods, knowledge-enhanced patient representations, graph-based retrieval systems, and KG-grounded medical LLMs.

\section{Method: multimodal instance--knowledge graphs}
\label{sec:method}
\textbf{Overview.} \graphname{} represents patient evidence and biomedical
knowledge within a single graph while keeping their node sets disjoint,
connected only by explicit alignment edges.
This separation is the central design decision.
Because each alignment edge is an individually addressable object rather than a
component of a fused representation, it can be retrieved, traced, and removed,
and the contribution of a given patient--relation correspondence to a downstream
prediction can be measured directly rather than inferred.
Modalities are accordingly not fused into a shared embedding, and the benefit comes from the correspondence rather than from added graph text: length-matched graph text with the correspondence disrupted reduces held-out ranking to near chance (\S\ref{sec:ranking-results}).

\subsection{Problem setting and graph representation}

\begin{figure}[htbp]
    \centering
    \includegraphics[width=\textwidth]{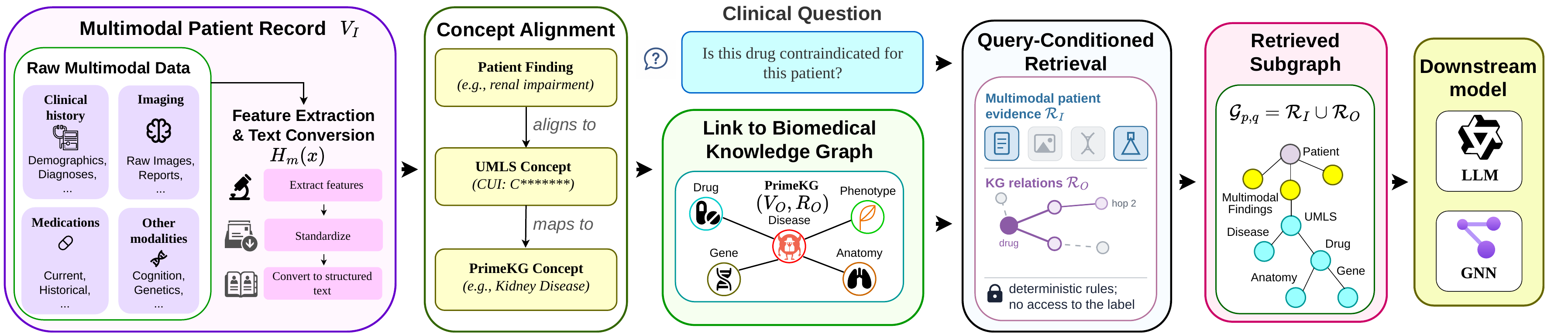}
    \caption{\textbf{Overview of \graphname{}.} Harmonizers convert each record into typed, provenance-linked observations; an aligner links them to PrimeKG concepts; query-conditioned retrieval supplies a subgraph to an LLM/GNN.}
    \label{fig:expanded-workflow}
\end{figure}

Let $\mathcal{D}=\{x_{p,e,m}\}$ be a multimodal corpus, where $p$ indexes an instance, $e$ an event/encounter, and $m\in\mathcal{M}$ a modality.
Let $\mathcal{B}=(V_O,E_O,R_O)$ be an external ontology or KG.
Given a query $q$ about instance $p$, the goal is to answer from a compact subgraph while retaining a trace to the supporting evidence in $\mathcal{D}$ or $\mathcal{B}$.
Figure~\ref{fig:expanded-workflow} summarizes the general architecture from multimodal inputs to query-conditioned model consumption.
We define a multimodal instance--knowledge graph as
\begin{equation*}
\mathcal{G}=(V,E,R,T,\tau,a,\pi),\qquad
V=V_I\cup V_O,\qquad
R=R_{\mathrm{proc}}\cup R_{\mathrm{align}}\cup R_{O}.
\end{equation*}
Here, $V_I$ and $V_O$ are the disjoint sets of instance and ontology nodes, $T$ is the set of node
types, each edge $(u,r,v)\in E$ carries a type $r\in R$, and the functions
$\tau$, $a$, and $\pi$ store node types, attributes, and provenance.
The partition of $R$ is what allows each layer to be traced or removed
independently: $R_{\mathrm{proc}}$ edges record how each observation was derived,
$R_{\mathrm{align}}$ edges link observations to concepts, and $R_{O}$ edges hold
the external knowledge.
Equivalently, $\mathcal{G}=\mathcal{G}_I\cup\mathcal{B}\cup\mathcal{G}_A$, where $\mathcal{G}_I$ is the instance graph and $\mathcal{G}_A$ contains alignment edges.
App.~\ref{app:formal} gives the fragment-level composition.

\subsection{Provenance-preserving multimodal construction}
MM-KG construction preserves the derivation path of every observation: each
observation node retains a pointer to its source record and to the time of
recording, so that any prediction can be traced to a specific measurement.
For every instance, we first build a local event graph:
\mbox{$
p \xrightarrow{\textsc{has-event}} e
\xrightarrow{\textsc{has-record}} d
\xrightarrow{\textsc{yields}} z$},
where $d$ is a source record and $z$ is a modality-derived observation.
The source pointer and timestamp are retained in $\pi$.
Each modality-specific harmonizer produces a graph fragment:
$
\mathcal{G}_{p,e,m}=H_m(x_{p,e,m};\phi_m),
%\quad
V_{p,e,m}=\{p,e,d,z_1,\ldots,z_k\}.
$
$H_m$ converts raw input into typed observations, edges, attributes, and provenance using fixed parameters $\phi_m$.
Examples include vocabulary lookup for codes, entity linking for text, vision models for imaging, or value-aware rules for measurements.
Dataset-specific harmonizers are summarized in \S\ref{sec:design} and detailed in Apps.~\ref{app:cohort}, \ref{app:imaging}, and~\ref{app:adnigraph}.
% For chest radiographs, the vision model converts finding scores into textual finding nodes that remain linked to the source image and later align to phenotype concepts.
% App.~\ref{app:imaging} gives the thresholding and calibration details.

Observations from different modalities remain distinct even when aligned to the same concept, preserving their source paths and supporting modality auditing and ablation.
All derived observations retain provenance, instance and ontology nodes remain disjoint, and harmonization, alignment, and retrieval are frozen before test labels are revealed (App.~\ref{app:formal}).

\subsection{Ontology alignment}
\label{sec:alignment}

Alignment assigns at most one \textsc{aligns-to} edge per observation, so deleting that edge detaches the observation from its concept, and from every relation of that concept, while leaving the rest of the graph unchanged.
For observation $z$, the aligner collects candidate concepts $C_m(z)\subseteq V_O$ through modality-specific mapping routes, such as exact identifiers, synonyms, and parent concepts, selects $c^*=\arg\max_{c\in C_m(z)}s_m(z,c)$, and adds $(z,\textsc{aligns-to},c^*)$ to $E$ only if $\Gamma_m(z,c^*)=1$.
Here $s_m$ ranks candidates by mapping route (exact identifier, then synonym, then guarded parent, then blocklisted name match) , and $\Gamma_m$ a type-and-route validity check; in our instantiation acceptance is route-based rather than thresholded.
Exact identifiers are preferred over synonym or parent routes, observations without a valid mapping remain on the instance side, and all mappings are frozen before evaluation.
App.~\ref{app:linking} gives complete routes and coverage statistics.

\subsection{Query-conditioned subgraph retrieval}

Serializing the full instance graph and ontology would produce an impractically large input, so the model receives only a subgraph conditioned on the question.
For query $q$ about instance $p$, we retrieve two typed components,
$\mathcal{G}_{p,q}=\mathcal{R}_{I}(\mathcal{G},p,q)\cup\mathcal{R}_{O}(\mathcal{G},q)$.
$\mathcal{R}_{I}$ returns the observations of $p$ available at prediction time, under fixed per-modality caps, and in our experiments does not depend on $q$; $\mathcal{R}_{O}$ returns the knowledge-graph relations of the entities in the question.
Both follow deterministic rules without access to the label, so the input grows with $\mathcal{G}_{p,q}$ rather than with $\mathcal{G}$.

Let $C_q\subseteq V_O$ be the entities linked from $q$, such as the candidate drugs a question asks about, and $R_q\subseteq R_O$ the relation types whitelisted for the question family.
For one-hop retrieval,
$
N_{R_q}(C_q)=
\{v\in V_O:\exists c\in C_q,\exists r\in R_q,
(c,r,v)\in E_O\},
$
and $\mathcal{R}_{O}$ returns the edges $(c,r,v)$ that define this set.
Multi-hop templates apply this operator recursively with a relation whitelist at each hop.
We call the relations $\mathcal{R}_{O}$ returns for one candidate that candidate's \emph{packet}.
The two components connect when a retrieved observation aligns to a concept in a packet, $z\xrightarrow{\textsc{aligns-to}}v$ with $(c,r,v)\in E_O$, $c\in C_q$, and $r\in R_q$; this path from the source observation through its aligned concept to the queried relation is the patient--relation correspondence that the factorial experimental design (\S \ref{sec:results-composite}) tests.

For set-membership questions, each packet contains the complete list of the requested relation type rather than a truncated list; \S\ref{sec:results-single} and App.~\ref{app:completeness} test this choice.
The subgraph is serialized as short natural-language statements grouped under source-channel headers (main experiments) or as typed triples (Table~\ref{tab:multimodel}).
App.~\ref{app:formal} records the correspondence path as an evidence certificate, which documents the evidence supplied to the model but not whether the graph or the answer is correct.

\subsection{Consumers and factorial evaluation}
\label{sec:factorial}

\graphname{} supports different consumers rather than defining a single predictor.
An LLM receives a serialization of $\mathcal{G}_{p,q}$ as its prompt, while a heterogeneous GNN propagates type-specific node features from frozen text and image encoders over the graph (App.~\ref{app:gnn}).
Outcome prediction has no query entities; there, the LLM receives the patient's record with the PrimeKG relations of that patient's own drugs and conditions, and the GNN receives the patient's graph (\S\ref{sec:results-single}).
% Our trained graph consumer uses two-layer heterogeneous GraphSAGE; App.~\ref{app:gnn} gives its architecture and training protocol.

We use a $2\times2$ experimental design in \S\ref{sec:results-composite} to test whether patient evidence and external KG relations interact.
All four conditions receive the question and the conventional patient record, $\mathcal{G}_{p,q}^{\mathrm{base}}$; B adds the patient modalities retrieved by $\mathcal{R}_I$, $\mathcal{G}_{p,q}^{\mathrm{mod}}$, C adds the packets retrieved by $\mathcal{R}_O$, $\mathcal{G}_{p,q}^{\mathrm{KG}}$, and D adds both.
Equivalently,
$
\mathcal{G}_{p,q}^{ij}
=\mathcal{G}_{p,q}^{\mathrm{base}}
\cup i\,\mathcal{G}_{p,q}^{\mathrm{mod}}
\cup j\,\mathcal{G}_{p,q}^{\mathrm{KG}},
\quad
(A,B,C,D)=(00,10,01,11).
$
Here, \(0\mathcal{G}=\emptyset\) and \(1\mathcal{G}=\mathcal{G}\).
For metric $M$, the interaction is
$
I_M=\big[M(D)-M(C)\big]-\big[M(B)-M(A)\big].
$
For metrics where higher is better, $I_M>0$ means that one source becomes more useful when the other is present.
Deleting the answer-bearing relation from a packet, swapping packets between candidates, and keeping node names without their relations test whether the interaction requires the correct correspondence rather than extra text or familiar entity names (\S\ref{sec:ranking-results}).

\section{Datasets and Implementation}
\label{sec:design}

We instantiate \graphname{} as independently constructed MIMIC and ADNI graphs linked to PrimeKG \citep{chandak2023primekg}, using the same provenance-preserving construction with dataset-specific extractors (Appendices~\ref{app:cohort}, \ref{app:imaging}, and~\ref{app:adnigraph}).
The MIMIC graph covers 51{,}844 patients, combining MIMIC-IV EHR events \citep{johnson2023mimiciv} with MIMIC-CXR radiographs and reports \citep{johnson2019mimiccxr,johnson2019mimiccxrjpg}: diagnoses, prescriptions, procedures, and laboratory measurements become provenance-linked observations, and BiomedCLIP \citep{zhang2023biomedclip} with CheXpert-calibrated thresholds \citep{irvin2019chexpert} converts radiographs into textual findings.
The ADNI graph covers 1{,}000 participants and 4{,}977 visits spanning imaging, medical history, cognition, genetics, and biospecimens \citep{veitch2022adni}; regional brain-MRI features from FastSurfer \citep{Henschel2020FastSurfer} are compared with normative models, their deviations become imaging findings, and direction-aware anchors to HPO phenotypes connect these findings to the other clinical domains.

We evaluate outcome prediction using 28{,}582 MIMIC ICU stays and 619 ADNI participants, biomedical fact-checking using 1{,}920 balanced questions, and patient-specific joins using 3{,}200 MIMIC items and the ADNI paired contraindication benchmark.
The ADNI benchmark contains 510 development items, a 114-item validation set and a 70-item test set; the latter provides the primary cross-cohort ADNI result.
The primary MIMIC experiments use Qwen3.5-35B-A3B and a two-layer heterogeneous GraphSAGE consumer, while the primary ADNI test evaluation uses Qwen3.5-9B with thinking disabled.
We report AUROC or balanced accuracy as appropriate, with paired-item bootstrap inference for MIMIC and participant-clustered inference for ADNI.
Further details on entity linking, evaluation settings, model implementation, and ADNI-specific evaluation details are provided in Appendices~\ref{app:linking}, \ref{app:evaluation}, \ref{app:gnn}, and \ref{app:adnieval}, respectively.

\section{Experiments and Results}
\label{sec:experiments-results}

We will evaluate \graphname{} on whether patient evidence and biomedical knowledge interact (\S\ref{sec:results-composite}), how \graphname{} compares with external methods (\S\ref{sec:external}), whether the interaction depends on the alignment edges (\S\ref{sec:ranking-results}), retrieval context efficiency (\S\ref{sec:retrieval-results}), and how each source performs alone (\S\ref{sec:results-single}).

\subsection{Aligned patient and knowledge evidence is superadditive}
\label{sec:results-composite}

Because instance and ontology nodes remain separately addressable, either evidence source can be withheld or supplied independently, which permits the $2{\times}2$ design of \S\ref{sec:factorial}: condition A receives neither source, B the patient modalities, C the KG relations, and D both.
Fused representations do not admit this design.
We call the two sources \emph{superadditive} when the improvement from supplying both exceeds the sum of the improvements from supplying each alone, that is, when the interaction $I_M$ of \S\ref{sec:factorial} is positive.

The MIMIC join benchmark targets questions for which such an interaction is expected: across all 28{,}582 MIMIC-IV ICU stays, 29.9\% exhibit a laboratory-evident organ state, (e.g., creatinine ${>}2.0$, bilirubin ${>}2.0$, platelets ${<}100$, or hemoglobin ${<}8$) that is absent from the coded diagnoses while PrimeKG lists a current medication as contraindicated for that state.
Items come from four question families with same-patient matched negatives and length-matched KG lists (App.~\ref{app:benchcons}).

\begin{table}[htbp]
\caption{Factorial results in both cohorts: balanced accuracy (chance 0.500) with AUROC in parentheses. \textbf{A}: conventional record; \textbf{B}: $+$patient modalities; \textbf{C}: $+$complete KG list; \textbf{D}: both. Interaction: $I_M$.}
\label{tab:composite}
\begin{center}
% \fontsize{8pt}{8pt}\selectfont
% \scriptsize
\setlength{\tabcolsep}{4.5pt}
\begin{tabular}{lccccc}
\toprule
Family & A & B & C & D & Interaction \\
\midrule
contra-report & 0.496 (.59) & 0.519 (.61) & 0.546 (.57) & \best{0.751 (.82)} & +0.182 ( +0.228) \\
contra-labs & 0.507 (.61) & 0.516 (.65) & 0.560 (.52) & \best{0.636 (.68)} & +0.068 ( +0.121) \\
mechanism & 0.524 (.52) & 0.510 (.55) & 0.526 (.53) & \best{0.616 (.67)} & +0.103 ( +0.118) \\
drug-effect & 0.475 (.37) & 0.449 (.31) & 0.494 (.49) & \best{0.600 (.88)} & +0.132 ( +0.463) \\
\midrule
All ($n{=}3{,}200$) & 0.501 (.52) & 0.498 (.51) & 0.532 (.53) & \best{0.651 (.72)} & +0.122 ( +0.209) \\
\midrule
\multicolumn{6}{l}{\emph{ADNI paired contraindication,
test set ($n{=}70$)}} \\
Qwen3.5-9B & 0.500 (.56) & 0.571 (.57) & 0.500 (.55) & \best{0.786 (.86)} & +0.214 ( +0.299) \\
Qwen2.5-7B-Instruct & 0.500 (.50) & 0.514 (.59) & 0.486 (.51) & \best{0.629 (.71)} & +0.129 ( +0.107) \\
\midrule
\multicolumn{6}{l}{\emph{ADNI paired contraindication, validation set ($n{=}114$)}} \\
Qwen3.5-9B & 0.500 (.56) & 0.535 (.55) & 0.500 (.55) & \best{0.746 (.86)} & +0.211 ( +0.319) \\
Qwen2.5-7B-Instruct & 0.500 (.49) & 0.509 (.56) & 0.500 (.48) & \best{0.561 (.68)} & +0.053 ( +0.129) \\
\bottomrule
\end{tabular}
\end{center}
\end{table}

Table~\ref{tab:composite} reports the factorial results on the 3{,}200 MIMIC items and the 70-item ADNI test set.
On MIMIC, supplying the patient modalities alone changes balanced accuracy by $-0.002$ and the KG alone by $+0.031$ ($p{<}0.001$), so additive effects would improve condition D by 0.029; it improves by 0.150, an interaction of $+0.122$ in balanced accuracy and $+0.209$ in AUROC ($p{<}0.001$).
The interaction is positive and significant in every family on both metrics, including the two-hop \emph{mechanism} questions, and remains $+0.179$ (95\% CI $[+0.160,+0.197]$) on the 2{,}400 items outside the drug-effect family, so the pooled estimate is not driven by that family's large effect.

\begin{figure}[t]
    \centering
    \includegraphics[width=0.55\textwidth]{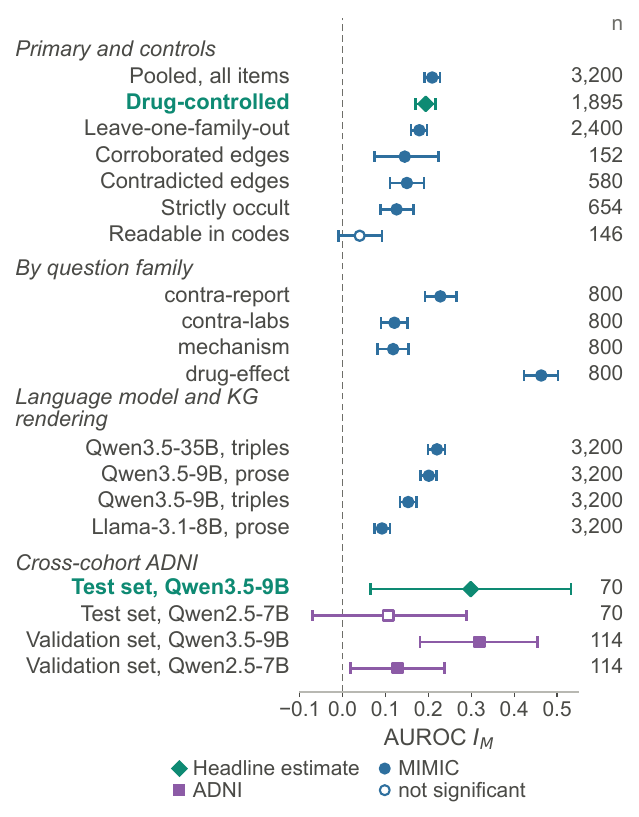}
    \caption{\textbf{Patient--KG interaction across question families, controls, language models, and cohorts:} AUROC $I_M$ with 95\% CI; marker area $\propto\sqrt{n}$ (Table~\ref{tab:interaction-robustness}).}
    \label{fig:interaction-forest}
\end{figure}

The lower blocks of Table~\ref{tab:composite} apply the same design to the independently constructed ADNI graph, using a paired contraindication benchmark that asks whether a candidate drug has a PrimeKG contraindication relation to a condition in the participant's medical history. Positive and negative items are paired across participants for the same drug and receive identical KG packets, so candidate identity alone cannot determine the label.
On the 70-item test set, scored once after the validation analysis and pipeline were frozen, the AUROC interaction is $+0.299$ (95\% CI $[+0.065,+0.532]$, $p=0.0135$), close to the $+0.319$ on the 114-item validation set, and it remains positive with Qwen2.5-7B-Instruct on both.
All four prespecified destructive controls fall below condition D after Holm correction, and removing the decisive contraindication edge alone reduces AUROC from 0.860 to 0.714 (App.~\ref{app:adnieval}).
Thus, while ADNI differs from MIMIC in disease domain, evidence channel, and language model, the interaction is still reproduced.

This interaction persists in all conducted control experiments (Figure~\ref{fig:interaction-forest}).
Restricting to the 1{,}895 items whose candidate drug appears in both label classes brings condition A to chance (AUROC 0.485) while the interaction remains $+0.194$ ($p{<}0.001$); it is positive across three LLMs and under prose or typed-triple rendering (App. Table~\ref{tab:multimodel}), and is preserved when answer-bearing edges are stratified by independent prescribing behavior (App.~\ref{app:validation}).
It is also specific to conditions documented outside the codes: when coded diagnoses already reveal the condition, C reaches 0.783 and the interaction disappears (App. Table~\ref{tab:mechanism}), so the interaction requires the laboratory and report evidence that codes do not contain.
Overall, neither patient evidence nor biomedical knowledge is sufficiently informative in isolation; modeling their interactions is required.

\subsection{\graphname{} outperforms existing methods on patient-specific reasoning}
\label{sec:external}

We next compare \graphname{} with appropriate existing methods on held-out five-candidate ranking: each item pairs one ICU stay with five administered drugs, one of which has an answer-bearing PrimeKG path, with decoys matched on prescription frequency, graph degree, packet length, and drug class ($n{=}1{,}244$ patients; chance Hits@1 0.200).
Condition D reaches Hits@1 0.575 (95\% CI $[0.548,0.603]$) against 0.293, 0.304, and 0.278 for A, B, and C, an interaction of $+0.287$ (95\% CI $[0.244,0.331]$, $p<0.001$).

Table~\ref{tab:ranking-external} compares \graphname{} with two external methods.
MindMap \citep{wen2024mindmap} +PrimeKG, scored with the same Qwen3.5 protocol, trails \graphname{} by 0.131 Hits@1 (95\% CI $[0.100,0.163]$, $p<0.001$).
An adapted GraphCare-BAT model \citep{jiang2024graphcare} trained on 19{,}085 labeled pairs is statistically indistinguishable from \graphname{} on the full split, but its score rises from 0.609 to 0.645 when the patient side of its input is removed ($p<0.001$), indicating that it ranks candidates by properties of the drug rather than by their relation to the patient.
On the identity-hard subset ($n=598$), where a candidate-only prior ranks the wrong drug and a formulary lookup cannot resolve the item, \graphname{} reaches 0.582 while both GraphCare variants fall to 0.375 and 0.381.
On ADNI, a deterministic PrimeKG lookup reaches AUROC 0.956, so that comparison measures recovery of an explicit patient--relation match rather than differences between methods (App. Table~\ref{tab:adni-external}); the five-candidate design could not be reproduced there because participant medication sets are too small to defeat candidate-identity shortcuts.
Overall, \graphname{} matches the strongest existing method where candidate identity suffices and exceeds every external method where it does not.

\begin{table}[t]
\caption{External methods on held-out five-candidate ranking (MIMIC; chance Hits@1 0.2). ``Identity-hard'': items a candidate-only prior ranks wrongly.}
\label{tab:ranking-external}
\centering
% \scriptsize
\setlength{\tabcolsep}{3pt}
\begin{tabular}{lccc}
\toprule
& \multicolumn{2}{c}{Full ($n{=}1{,}244$)} & Id.-hard ($n{=}598$) \\
\cmidrule(r){2-3}\cmidrule(l){4-4}
Method & Hits@1 & MRR & Hits@1 \\
\midrule
\graphname{} (condition D) & \underline{0.575} & \underline{0.738} & \best{0.582} \\
GraphCare-BAT adapted & \best{0.609} & \best{0.764} & 0.375 \\
\quad \emph{(patient side removed)} & \emph{0.645} & \emph{0.787} & \emph{0.381} \\
MindMap-PrimeKG & 0.444 & 0.646 & \underline{0.448} \\
\bottomrule
\end{tabular}
\end{table}

\subsection{Patient--KG interaction requires correct alignment, not graph text}
\label{sec:ranking-results}

Because retrieval returns each relation as a separately addressable object, the patient--relation correspondence can be disrupted while the patient evidence and nearly all of the graph text stay in place.
We apply three such disruptions to condition D on the five-candidate MIMIC ranking benchmark of \S\ref{sec:external}: deleting the single answer-bearing relation from the answer candidate's packet, replacing the answer candidate's packet (the relations retrieved for it) with a length-matched packet from another candidate, and shuffling packets across the five candidates.
All three fall below condition D ($p<0.001$; Table~\ref{tab:disruption}): deleting the edge returns Hits@1 to the level of condition A, and shuffling drives it below the 0.200 chance level.
Because the decoys are matched on prescription frequency, graph degree, packet length, and drug class, the drop reflects the loss of the correspondence rather than of graph text.
The ADNI 70-item test set shows the same ordering, with all four prespecified disruptions below condition D (after Holm correction): removing the decisive edge lowers AUROC from 0.860 to 0.714, and keeping node names while removing their relations returns it to the level of condition A.
\graphname{} also cites the evidence it uses: on the ADNI validation set, all 85 attempted KG citations referred to relations in the supplied packet, and after the decisive edge was removed the model never cited it (App.~\ref{app:adnieval}).

\begin{table}[t]
\caption{\textbf{Correspondence disruption.} MIMIC: five-candidate ranking ($n{=}1{,}244$); ADNI: test set ($n{=}70$). $^{a}$on MIMIC: another candidate's length-matched packet; on ADNI: degree- and relation-matched null packet.}
\label{tab:disruption}
\centering
% \scriptsize
\setlength{\tabcolsep}{3pt}
\begin{tabular}{lcc}
\toprule
& MIMIC (Hits@1) & ADNI (AUROC) \\
\midrule
Correct alignment (D) & \best{0.575} & \best{0.860} \\
Answer-bearing edge removed & 0.299 & 0.714 \\
Matched control packet$^{a}$ & 0.237 & 0.548 \\
Shuffled packets & 0.185 & 0.531 \\
Node names without relations & --- & 0.558 \\
\midrule
No knowledge (A) & 0.293 & 0.558 \\
Chance & 0.200 & 0.500 \\
\bottomrule
\end{tabular}
\end{table}

The patient-side contribution localizes in the same way.
On MIMIC, removing laboratory values or report findings reduces Hits@1 by 0.148 and 0.123 ($p{<}0.001$), whereas removing imaging findings or coded history has no measurable effect (App.~\ref{app:mimic-ranking}); on the ADNI validation set, medical history alone with the KG reaches AUROC 0.831 against 0.853 for the full record, and it is the only patient block whose enablement estimate ($+0.294$) exceeds the prespecified minimum detectable interaction of 0.137 (App.~\ref{app:adnieval}).
The interaction therefore depends on the alignment edges: mismatched packets return both cohorts to the no-knowledge level, and in each cohort the interaction arises from the evidence block that contains the condition the question concerns.

\subsection{Query-conditioned retrieval recovers correspondence with less context}
\label{sec:retrieval-results}

\begin{figure}[t]
    \centering
    \includegraphics[width=0.7\textwidth]{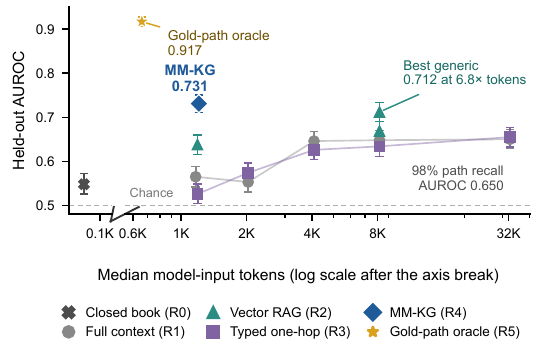}
    \caption{\textbf{Retrieval accuracy against context cost} on the held-out MIMIC split: test AUROC (95\% CI) vs.\ median input tokens.}
    \label{fig:retrieval-cost}
\end{figure}

We next measure the cost of retrieving the correspondence on the patient-disjoint MIMIC test split (2{,}398 items, 1{,}079 patients), scored once after all policies were frozen (App.~\ref{app:mimic-retrieval}).
\graphname{} reaches AUROC 0.731 with 1{,}206 input tokens, $6.8\times$ fewer than the 8{,}157 used by the strongest generic policy, SapBERT vector retrieval (0.712, $p{=}0.01$); the remaining policies score 0.062--0.204 lower (Figure~\ref{fig:retrieval-cost}), and when capped at \graphname{}'s budget they recover the answer path for at most 26\% of positive items and fall to 0.526--0.638.
On the drug-balanced subset, whose candidate drugs appear in both label classes, the two leading policies tie (0.695 against 0.693) while the token gap remains ($6.7\times$), so the consistent advantage is context efficiency.
MindMap-PrimeKG reaches 0.686 and recovers a complete answer path for 72.6\% of positive items, against 99.7\% for \graphname{}.

Retrieval can fail by missing the answer path or by retrieving it without the model using it.
\graphname{} misses the path on only 4 of 1{,}199 positive items and resolves more positives than any other deployable policy: 753, against 714 for the 32K full context, which retrieves 98.2\% of paths but loses 463 positives with the path present (App. Figure~\ref{fig:failure-decomposition-full}).
On ADNI, varying path length from one to three hops showcases both failure modes: a static relation packet falls below closed-book answering despite usually containing the gold path, and interactive graph querying is accurate at one hop but drops to 0.670 and 0.590 balanced accuracy at two and three hops (App.~\ref{app:adniretrieval}).
Query-conditioned retrieval thus supplies the required correspondence at a fraction of the context, with fewer combined failures than any generic policy.

\subsection{MM-KG's multimodal patient representations for outcome prediction}
\label{sec:results-single}

\begin{table}[t]
\caption{\textbf{ICU mortality ablations:} $\Delta$AUROC when each channel is removed (ablated $-$ full). LLM: 28{,}582 stays; GNN: 5{,}676-stay test split. $^{***}p{<}0.001$, $^{*}p{<}0.05$.}
\label{tab:prediction}
\centering
% \scriptsize
\setlength{\tabcolsep}{4pt}
\begin{tabular}{@{}lrr@{}}
\toprule
Channel & LLM (zero-shot) & GNN (trained) \\
& ($\Delta$)AUROC & ($\Delta$)AUROC \\
\midrule
Full model & $0.733$ & $0.804$ \\
\midrule
\quad $-$ Laboratory & \best{$-0.049^{***}$} & \best{$-0.085^{***}$}  \\
\quad $-$ Prior-visit history & $-0.024^{***}$ & $-0.005^{***}$ \\
\quad $-$ Medications & $-0.011^{***}$ & $-0.001$ n.s. \\
\quad $-$ Radiology report & $-0.011^{***}$ & $+0.001$ n.s. \\
\quad $-$ Imaging findings & $-0.002$ n.s. & $-0.008^{*}$ \\
\quad $-$ PrimeKG & $+0.002$ n.s. & $-0.002$ n.s. \\
\bottomrule
\end{tabular}
\end{table}

Finally, we examine each source on tasks that do not require their combination.
For ICU mortality, the zero-shot LLM reaches AUROC 0.733 with the full \graphname{} prompt, against 0.570 with demographics alone, and the trained GNN reaches 0.804, against 0.778 for a laboratory-only model; laboratory values contribute most for both (Table~\ref{tab:prediction}).
The GNN also outperforms logistic regression on 30-day readmission (0.644 against 0.548; App.~\ref{app:gnn}), the harmonizer's structured imaging findings exceed the radiologist's verbatim prose by $+0.025$ AUROC (App.~\ref{app:imaging}), and in ADNI, adding biomarkers and imaging to clinical history improves prognosis balanced accuracy from 0.595 to 0.652 (App.~\ref{app:adnieval}).
Removing PrimeKG leaves mortality AUROC unchanged for both consumers, within run-to-run noise (App.~\ref{app:noise}), consistent with \S\ref{sec:results-composite}: the knowledge graph contributes when a question requires the patient--relation correspondence.

Conversely, when a question concerns a relation directly, complete PrimeKG edges improve balanced accuracy on 1{,}920 pharmacological questions in all eight relation families, by $+0.063$ to $+0.420$, whereas entity names, length-matched incorrect neighborhoods, and corrupted names do not (App. Table~\ref{tab:audit}); because the model follows behaviorally corroborated and contradicted edges alike (App.~\ref{app:validation}), traceability to an edge does not establish clinical correctness.
Each source is therefore informative on the task suited to it, and interaction (\S\ref{sec:results-composite}) arises when a question requires both.

\section{Conclusion}

We introduced \graphname{}, which joins multimodal patient evidence to a biomedical knowledge graph through explicit, provenance-preserving alignment edges.
In two independently constructed cohorts, the two sources are individually uninformative on questions that require both but superadditive together, and replacing the aligned relations with mismatched ones returns performance to the no-knowledge level.
Query-conditioned retrieval recovers these links with a fraction of the context of generic retrieval, and \graphname{} outperforms external graph methods on the items that require consulting the patient.
Knowledge graphs thus help clinical LLMs not as background context but as explicit links between a patient's state and the relation a question requires.

\bibliographystyle{style/icml2025}
\bibliography{999_reference}

\begin{thebibliography}{34}
\providecommand{\natexlab}[1]{#1}
\providecommand{\url}[1]{\texttt{#1}}
\expandafter\ifx\csname urlstyle\endcsname\relax
  \providecommand{\doi}[1]{doi: #1}\else
  \providecommand{\doi}{doi: \begingroup \urlstyle{rm}\Url}\fi

\bibitem[Bodenreider(2004)]{bodenreider2004umls}
Bodenreider, O.
\newblock The {Unified Medical Language System} ({UMLS}): integrating biomedical terminology.
\newblock \emph{Nucleic Acids Research}, 32\penalty0 (suppl\_1):\penalty0 D267--D270, 2004.

\bibitem[Burger et~al.(2023)Burger, R{\"a}tsch, and Kuznetsova]{burger2023mmugl}
Burger, M., R{\"a}tsch, G., and Kuznetsova, R.
\newblock Multi-modal graph learning over {UMLS} knowledge graphs.
\newblock In \emph{Proceedings of the 3rd Machine Learning for Health Symposium}, volume 225 of \emph{Proceedings of Machine Learning Research}, pp.\  52--81, 2023.
\newblock URL \url{https://proceedings.mlr.press/v225/burger23a.html}.

\bibitem[Chandak et~al.(2023)Chandak, Huang, and Zitnik]{chandak2023primekg}
Chandak, P., Huang, K., and Zitnik, M.
\newblock Building a knowledge graph to enable precision medicine.
\newblock \emph{Scientific Data}, 10\penalty0 (1):\penalty0 67, 2023.

\bibitem[Chen et~al.(2024)Chen, Zhang, Fang, Geng, Guo, Chen, Li, Zhang, Chen, Zhu, Li, Liu, Pan, Zhang, and Chen]{chen2024mmkgsurvey}
Chen, Z., Zhang, Y., Fang, Y., Geng, Y., Guo, L., Chen, X., Li, Q., Zhang, W., Chen, J., Zhu, Y., Li, J., Liu, X., Pan, J.~Z., Zhang, N., and Chen, H.
\newblock Knowledge graphs meet multi-modal learning: A comprehensive survey.
\newblock \emph{arXiv preprint arXiv:2402.05391}, 2024.
\newblock \doi{10.48550/arXiv.2402.05391}.

\bibitem[Choi et~al.(2016)Choi, Bahadori, Sun, Kulas, Schuetz, and Stewart]{choi2016retain}
Choi, E., Bahadori, M.~T., Sun, J., Kulas, J., Schuetz, A., and Stewart, W.
\newblock {RETAIN}: An interpretable predictive model for healthcare using reverse time attention mechanism.
\newblock In \emph{Advances in Neural Information Processing Systems (NeurIPS)}, volume~29, 2016.

\bibitem[Choi et~al.(2017)Choi, Bahadori, Song, Stewart, and Sun]{choi2017gram}
Choi, E., Bahadori, M.~T., Song, L., Stewart, W.~F., and Sun, J.
\newblock {GRAM}: Graph-based attention model for healthcare representation learning.
\newblock In \emph{Proceedings of the 23rd ACM SIGKDD International Conference on Knowledge Discovery and Data Mining}, pp.\  787--795, 2017.

\bibitem[Elsharief et~al.(2025)Elsharief, Shurrab, Jorf, Lopez, Geras, and Shamout]{elsharief2025medmod}
Elsharief, S., Shurrab, S., Jorf, B.~A., Lopez, L. J.~L., Geras, K.~J., and Shamout, F.~E.
\newblock {MedMod}: Multimodal benchmark for medical prediction tasks with electronic health records and chest x-ray scans.
\newblock In \emph{Proceedings of the Sixth Conference on Health, Inference, and Learning}, volume 287 of \emph{Proceedings of Machine Learning Research}, pp.\  781--803. PMLR, 2025.
\newblock URL \url{https://proceedings.mlr.press/v287/elsharief25a.html}.

\bibitem[Gao et~al.(2023)Gao, Li, Caskey, Dligach, Miller, Churpek, and Afshar]{gao2025drknows}
Gao, Y., Li, R., Caskey, J., Dligach, D., Miller, T., Churpek, M.~M., and Afshar, M.
\newblock Leveraging a medical knowledge graph into large language models for diagnosis prediction.
\newblock \emph{arXiv preprint arXiv:2308.14321}, 2023.

\bibitem[Hayat et~al.(2022)Hayat, Geras, and Shamout]{hayat2022medfuse}
Hayat, N., Geras, K.~J., and Shamout, F.~E.
\newblock {MedFuse}: Multi-modal fusion with clinical time-series data and chest x-ray images.
\newblock In \emph{Proceedings of the 7th Machine Learning for Healthcare Conference}, volume 182 of \emph{Proceedings of Machine Learning Research}, pp.\  479--503, 2022.
\newblock URL \url{https://proceedings.mlr.press/v182/hayat22a.html}.

\bibitem[Henschel et~al.(2020)Henschel, Conjeti, Estrada, Diers, Fischl, and Reuter]{Henschel2020FastSurfer}
Henschel, L., Conjeti, S., Estrada, S., Diers, K., Fischl, B., and Reuter, M.
\newblock {FastSurfer} -- a fast and accurate deep learning based neuroimaging pipeline.
\newblock \emph{NeuroImage}, 219:\penalty0 117012, 2020.
\newblock \doi{10.1016/j.neuroimage.2020.117012}.
\newblock URL \url{https://doi.org/10.1016/j.neuroimage.2020.117012}.

\bibitem[Irvin et~al.(2019)Irvin, Rajpurkar, Ko, Yu, Ciurea-Ilcus, Chute, Marklund, Haghgoo, Ball, Shpanskaya, Seekins, Mong, Halabi, Sandberg, Jones, Larson, Langlotz, Patel, Lungren, and Ng]{irvin2019chexpert}
Irvin, J., Rajpurkar, P., Ko, M., Yu, Y., Ciurea-Ilcus, S., Chute, C., Marklund, H., Haghgoo, B., Ball, R., Shpanskaya, K., Seekins, J., Mong, D.~A., Halabi, S.~S., Sandberg, J.~K., Jones, R., Larson, D.~B., Langlotz, C.~P., Patel, B.~N., Lungren, M.~P., and Ng, A.~Y.
\newblock {CheXpert}: A large chest radiograph dataset with uncertainty labels and expert comparison.
\newblock In \emph{Proceedings of the AAAI Conference on Artificial Intelligence}, volume~33, pp.\  590--597, 2019.

\bibitem[Jiang et~al.(2024)Jiang, Xiao, Cross, and Sun]{jiang2024graphcare}
Jiang, P., Xiao, C., Cross, A., and Sun, J.
\newblock {GraphCare}: Enhancing healthcare predictions with personalized knowledge graphs.
\newblock In \emph{International Conference on Learning Representations (ICLR)}, 2024.

\bibitem[Jiang et~al.(2025)Jiang, Xiao, Jiang, Bhatia, Kass-Hout, Sun, and Han]{jiang2025kare}
Jiang, P., Xiao, C., Jiang, M., Bhatia, P., Kass-Hout, T., Sun, J., and Han, J.
\newblock Reasoning-enhanced healthcare predictions with knowledge graph community retrieval.
\newblock In \emph{International Conference on Learning Representations}, 2025.
\newblock URL \url{https://proceedings.iclr.cc/paper_files/paper/2025/hash/cb5a3b4589c1a16f14740396625cbfc8-Abstract-Conference.html}.

\bibitem[Johnson et~al.(2019{\natexlab{a}})Johnson, Pollard, Berkowitz, Greenbaum, Lungren, Deng, Mark, and Horng]{johnson2019mimiccxr}
Johnson, A.~E., Pollard, T.~J., Berkowitz, S.~J., Greenbaum, N.~R., Lungren, M.~P., Deng, C.-y., Mark, R.~G., and Horng, S.
\newblock {MIMIC-CXR}, a de-identified publicly available database of chest radiographs with free-text reports.
\newblock \emph{Scientific data}, 6\penalty0 (1):\penalty0 317, 2019{\natexlab{a}}.

\bibitem[Johnson et~al.(2019{\natexlab{b}})Johnson, Pollard, Greenbaum, Lungren, Deng, Peng, Lu, Mark, Berkowitz, and Horng]{johnson2019mimiccxrjpg}
Johnson, A.~E., Pollard, T.~J., Greenbaum, N.~R., Lungren, M.~P., Deng, C.-y., Peng, Y., Lu, Z., Mark, R.~G., Berkowitz, S.~J., and Horng, S.
\newblock {MIMIC-CXR-JPG}, a large publicly available database of labeled chest radiographs.
\newblock \emph{arXiv preprint arXiv:1901.07042}, 2019{\natexlab{b}}.

\bibitem[Johnson et~al.(2023)Johnson, Bulgarelli, Shen, Gayles, Shammout, Horng, Pollard, Hao, Moody, Gow, Lehman, Celi, and Mark]{johnson2023mimiciv}
Johnson, A.~E., Bulgarelli, L., Shen, L., Gayles, A., Shammout, A., Horng, S., Pollard, T.~J., Hao, S., Moody, B., Gow, B., Lehman, L.-w.~H., Celi, L.~A., and Mark, R.~G.
\newblock {MIMIC-IV}, a freely accessible electronic health record dataset.
\newblock \emph{Scientific Data}, 10\penalty0 (1):\penalty0 1, 2023.

\bibitem[Johnson et~al.(2026)Johnson, Gottlieb, Shaham, Eisen, Waxman, Devons-Sberro, Ginder, Hong, Sayeed, Su, Reis, Balicer, Dagan, and Zitnik]{johnson2024clinvec}
Johnson, R., Gottlieb, U., Shaham, G., Eisen, L., Waxman, J., Devons-Sberro, S., Ginder, C.~R., Hong, P., Sayeed, R., Su, X., Reis, B.~Y., Balicer, R.~D., Dagan, N., and Zitnik, M.
\newblock Embeddings of clinical codes enable knowledge-grounded {AI} in medicine.
\newblock \emph{npj Digital Medicine}, 2026.
\newblock \doi{10.1038/s41746-026-02664-9}.

\bibitem[Kwon et~al.(2023)Kwon, Li, Zhuang, Sheng, Zheng, Yu, Gonzalez, Zhang, and Stoica]{kwon2023vllm}
Kwon, W., Li, Z., Zhuang, S., Sheng, Y., Zheng, L., Yu, C.~H., Gonzalez, J.~E., Zhang, H., and Stoica, I.
\newblock Efficient memory management for large language model serving with {PagedAttention}.
\newblock In \emph{Proceedings of the 29th Symposium on Operating Systems Principles (SOSP)}, 2023.

\bibitem[Lewis et~al.(2020)Lewis, Perez, Piktus, Petroni, Karpukhin, Goyal, K{\"u}ttler, Lewis, Yih, Rockt{\"a}schel, Riedel, and Kiela]{lewis2020rag}
Lewis, P., Perez, E., Piktus, A., Petroni, F., Karpukhin, V., Goyal, N., K{\"u}ttler, H., Lewis, M., Yih, W.-t., Rockt{\"a}schel, T., Riedel, S., and Kiela, D.
\newblock Retrieval-augmented generation for knowledge-intensive {NLP} tasks.
\newblock In \emph{Advances in Neural Information Processing Systems (NeurIPS)}, volume~33, pp.\  9459--9474, 2020.

\bibitem[Li et~al.(2025)Li, Dai, Chen, Du, Yao, Zhang, and Wang]{li2025rad}
Li, H., Dai, T., Chen, Z., Du, S., Yao, J., Zhang, Y., and Wang, Y.
\newblock {RAD}: Towards trustworthy retrieval-augmented multi-modal clinical diagnosis.
\newblock In \emph{Advances in Neural Information Processing Systems}, volume~38, 2025.
\newblock \doi{10.52202/085713-4109}.
\newblock URL \url{https://papers.nips.cc/paper_files/paper/2025/hash/b28ca52a43881fd87196c71145f30092-Abstract-Conference.html}.

\bibitem[Park et~al.(2022)Park, Bae, Kim, Kim, and Choi]{park2022medgtx}
Park, S., Bae, S., Kim, J., Kim, T., and Choi, E.
\newblock Graph-text multi-modal pre-training for medical representation learning.
\newblock In \emph{Proceedings of the Conference on Health, Inference, and Learning}, volume 174 of \emph{Proceedings of Machine Learning Research}, pp.\  261--281. PMLR, 2022.
\newblock URL \url{https://proceedings.mlr.press/v174/park22a.html}.

\bibitem[Stear et~al.(2024)Stear, Mohseni~Ahooyi, Simmons, et~al.]{stear2024petagraph}
Stear, B.~J., Mohseni~Ahooyi, T., Simmons, J.~A., et~al.
\newblock Petagraph: A large-scale unifying knowledge graph framework for integrating biomolecular and biomedical data.
\newblock \emph{Scientific Data}, 11:\penalty0 1338, 2024.
\newblock \doi{10.1038/s41597-024-04070-w}.

\bibitem[Su et~al.(2025)Su, Messica, Huang, Johnson, Fesser, Gao, Sahneh, and Zitnik]{su2025medtok}
Su, X., Messica, S., Huang, Y., Johnson, R., Fesser, L., Gao, S., Sahneh, F., and Zitnik, M.
\newblock Multimodal medical code tokenizer.
\newblock In \emph{Proceedings of the 42nd International Conference on Machine Learning}, volume 267 of \emph{Proceedings of Machine Learning Research}, pp.\  57030--57045. PMLR, 2025.
\newblock URL \url{https://proceedings.mlr.press/v267/su25b.html}.

\bibitem[Veitch et~al.(2022)Veitch, Weiner, Aisen, et~al.]{veitch2022adni}
Veitch, D.~P., Weiner, M.~W., Aisen, P.~S., et~al.
\newblock Using the {Alzheimer's Disease Neuroimaging Initiative} to improve early detection, diagnosis, and treatment of {Alzheimer's} disease.
\newblock \emph{Alzheimer's \& Dementia}, 18\penalty0 (4):\penalty0 824--857, 2022.
\newblock \doi{10.1002/alz.12422}.

\bibitem[Wen et~al.(2024)Wen, Wang, and Sun]{wen2024mindmap}
Wen, Y., Wang, Z., and Sun, J.
\newblock {MindMap}: Knowledge graph prompting sparks graph of thoughts in large language models.
\newblock In \emph{Proceedings of the 62nd Annual Meeting of the Association for Computational Linguistics (ACL)}, 2024.

\bibitem[Wu et~al.(2025)Wu, Zhu, Qi, Chen, Xu, Menolascina, Jin, and Grau]{wu2024medgraphrag}
Wu, J., Zhu, J., Qi, Y., Chen, J., Xu, M., Menolascina, F., Jin, Y., and Grau, V.
\newblock Medical graph {RAG}: Evidence-based medical large language model via graph retrieval-augmented generation.
\newblock In \emph{Proceedings of the 63rd Annual Meeting of the Association for Computational Linguistics (Volume 1: Long Papers)}, pp.\  28443--28467, Vienna, Austria, 2025. Association for Computational Linguistics.
\newblock \doi{10.18653/v1/2025.acl-long.1381}.
\newblock URL \url{https://aclanthology.org/2025.acl-long.1381/}.

\bibitem[Wu et~al.(2024)Wu, Dadu, Tustison, Avants, Nalls, Sun, and Faghri]{wu2024muse}
Wu, Z., Dadu, A., Tustison, N., Avants, B., Nalls, M., Sun, J., and Faghri, F.
\newblock Multimodal patient representation learning with missing modalities and labels.
\newblock In \emph{The Twelfth International Conference on Learning Representations}, 2024.
\newblock URL \url{https://openreview.net/forum?id=Je5SHCKpPa}.

\bibitem[Xu et~al.(2024)Xu, Shi, Yu, Zhuang, Jin, Wang, Ho, and Yang]{xu2024ramehr}
Xu, R., Shi, W., Yu, Y., Zhuang, Y., Jin, B., Wang, M.~D., Ho, J.~C., and Yang, C.
\newblock {RAM-EHR}: Retrieval augmentation meets clinical predictions on electronic health records.
\newblock In \emph{Proceedings of the 62nd Annual Meeting of the Association for Computational Linguistics (Volume 2: Short Papers)}, pp.\  754--765, 2024.
\newblock \doi{10.18653/v1/2024.acl-short.68}.
\newblock URL \url{https://aclanthology.org/2024.acl-short.68/}.

\bibitem[Yao et~al.(2024{\natexlab{a}})Yao, Liu, Yin, Cheung, and Qin]{yao2024ddlcxr}
Yao, W., Liu, C., Yin, K., Cheung, W.~K., and Qin, J.
\newblock Addressing asynchronicity in clinical multimodal fusion via individualized chest x-ray generation.
\newblock In \emph{Advances in Neural Information Processing Systems}, volume~37, 2024{\natexlab{a}}.
\newblock \doi{10.52202/079017-0913}.

\bibitem[Yao et~al.(2024{\natexlab{b}})Yao, Yin, Cheung, Liu, and Qin]{yao2024drfuse}
Yao, W., Yin, K., Cheung, W.~K., Liu, J., and Qin, J.
\newblock {DrFuse}: Learning disentangled representation for clinical multi-modal fusion with missing modality and modal inconsistency.
\newblock In \emph{Proceedings of the AAAI Conference on Artificial Intelligence}, volume~38, pp.\  16416--16424, 2024{\natexlab{b}}.
\newblock \doi{10.1609/aaai.v38i15.29578}.

\bibitem[Zhang et~al.(2024)Zhang, Nagesh, Shyani, and Mishra]{zhang2024dossier}
Zhang, H., Nagesh, S., Shyani, M., and Mishra, N.
\newblock {DOSSIER}: Fact checking in electronic health records while preserving patient privacy.
\newblock In \emph{Proceedings of the 9th Machine Learning for Healthcare Conference}, volume 252 of \emph{Proceedings of Machine Learning Research}. PMLR, 2024.
\newblock URL \url{https://proceedings.mlr.press/v252/zhang24a.html}.

\bibitem[Zhang et~al.(2023)Zhang, Xu, Usuyama, Xu, Bagga, Tinn, Preston, Rao, Wei, Valluri, Wong, Tupini, Wang, Mazzola, Shukla, Liden, Gao, Lungren, Naumann, Wang, and Poon]{zhang2023biomedclip}
Zhang, S., Xu, Y., Usuyama, N., Xu, H., Bagga, J., Tinn, R., Preston, S., Rao, R., Wei, M., Valluri, N., Wong, C., Tupini, A., Wang, Y., Mazzola, M., Shukla, S., Liden, L., Gao, J., Lungren, M.~P., Naumann, T., Wang, S., and Poon, H.
\newblock {BiomedCLIP}: a multimodal biomedical foundation model pretrained from fifteen million scientific image-text pairs.
\newblock \emph{arXiv preprint arXiv:2303.00915}, 2023.

\bibitem[Zhou et~al.(2026)Zhou, Huang, Cole, Britton, Yin, Wolber, and Li]{zhou2026faith}
Zhou, S., Huang, M., Cole, J., Britton, C., Yin, M., Wolber, J., and Li, K.
\newblock Assessing automated fact-checking for medical {LLM} responses with knowledge graphs.
\newblock In \emph{Proceedings of the AAAI Conference on Artificial Intelligence}, volume~40, pp.\  39646--39656, 2026.
\newblock \doi{10.1609/aaai.v40i46.41317}.

\bibitem[Zuo et~al.(2025)Zuo, Jiang, Mo, and Li{\`o}]{zuo2025kg4diagnosis}
Zuo, K., Jiang, Y., Mo, F., and Li{\`o}, P.
\newblock {KG4Diagnosis}: A hierarchical multi-agent {LLM} framework with knowledge graph enhancement for medical diagnosis.
\newblock In \emph{Proceedings of The First AAAI Bridge Program on AI for Medicine and Healthcare}, volume 281 of \emph{Proceedings of Machine Learning Research}, pp.\  195--204, 2025.
\newblock URL \url{https://proceedings.mlr.press/v281/zuo25a.html}.

\end{thebibliography}

\clearpage
\appendix
\FloatBarrier
\section{Additional formal details}
\label{app:formal}

\subsection{Fragment-level graph composition}

Let $\mathcal{G}_{A}=(V_I\cup V_O,E_{\mathrm{align}})$ denote the alignment graph.
The complete graph can be written as
\begin{equation}
\mathcal{G}
=\mathcal{G}_{I}\cup\mathcal{B}\cup\mathcal{G}_{A},
\qquad
\mathcal{G}_{I}=\bigoplus_{p,e,m}\mathcal{G}_{p,e,m},
\label{eq:compose}
\end{equation}
where $\mathcal{G}_{p,e,m}$ is one modality fragment and $\oplus$ joins fragments through shared instance and event identifiers.
This decomposition allows each component to be stored, retrieved, or removed without changing node identities in the other components.
Formally, $\tau:V\rightarrow T$ assigns node types, $a(v)$ stores node attributes, and $\pi:V\cup E\rightarrow\mathcal{P}$ maps nodes and edges to provenance records.
Observations with independent identities become nodes, whereas dense measurements meaningful only within an event may remain event attributes.

\subsection{Construction invariants}

The representation enforces three invariants.
First, every derived observation has a source path:
\begin{equation}
\forall z\in V_I^{\mathrm{obs}},\quad
\exists(p,e,d): 
p\xrightarrow{}e\xrightarrow{}d\xrightarrow{}z,
\quad
\pi(z)=(\text{source},\text{id},\text{time}).
\label{eq:provenance}
\end{equation}
Second, instance observations and ontology concepts have separate identities,
$V_I\cap V_O=\emptyset$, even when they denote the same clinical idea.
Third, harmonization, alignment, and retrieval are frozen before test labels are revealed.
These invariants preserve source identity and prevent label-conditioned graph construction.

\subsection{Evidence-trace certificate}

A retrieved answer can include the evidence certificate
\begin{equation}
\Xi_{p,q}=(P_I,E_A,P_O),
\label{eq:certificate}
\end{equation}
where $P_I$ traces the instance to an observation, $E_A$ is its alignment edge, and $P_O$ is the domain path used for the answer.
The certificate is valid only if every edge exists in the frozen graph, every source predates the prediction cutoff, and every domain relation belongs to the query whitelist.
It records the evidence given to the model but does not establish that the graph or answer is correct.

\FloatBarrier
\section{Cohort construction and flow}
\label{app:cohort}
\graphname{} uses MIMIC-IV v3.1 and MIMIC-CXR-JPG v2.1.0.
The following table traces the flow from the raw datasets to each evaluation cohort.
\begin{center}
\small
\begin{tabular}{lrr}
\toprule
Stage & Patients & Units \\
\midrule
MIMIC-IV v3.1 (hosp) & 364{,}627 & 546{,}028 admissions \\
MIMIC-CXR-JPG v2.1.0 & 65{,}379 & 227{,}835 studies \\
Present in both datasets & 61{,}868 & --- \\
$+$ $\geq$1 admission, $\geq$1 coded dx (= \graphname{}) & 51{,}844 & 219{,}720 adm.; 204{,}016 studies \\
\midrule
ICU stays in \graphname{} (prediction and join-task base) & 18{,}396 & 28{,}582 stays \\
\quad of which lab-occult contraindication carriers & --- & 8{,}555 stays (29.9\%) \\
\quad Join-task items (400 matched pairs $\times$ 4 families) & --- & 3{,}200 items \\
Fact-checking pair cohort (Table~\ref{tab:audit} source) & 47{,}100 & 183{,}950 admissions \\
\bottomrule
\end{tabular}
\end{center}

\begin{table}[htbp]
\caption{MIMIC--PrimeKG graph composition.
The two sides are connected by 13{,}352 typed alignment edges.}
\label{tab:graph}
\centering
\footnotesize
\setlength{\tabcolsep}{4.5pt}
\begin{tabular}{lrlr}
\toprule
Patient-side type & Nodes & PrimeKG type & Nodes \\
\midrule
Patient & 51{,}844 & Disease & 17{,}080 \\
Admission & 219{,}720 & Effect/phenotype & 15{,}311 \\
CXR study & 204{,}016 & Drug & 7{,}957 \\
Finding & 28{,}714 & Exposure & 818 \\
Procedure & 9{,}348 & & \\
Drug / diagnosis & 5{,}908 / 84 & & \\
\midrule
\multicolumn{4}{l}{Total: 560{,}800 nodes and 14{,}407{,}417 edges.} \\
\bottomrule
\end{tabular}
\end{table}

``No sampling'' means that every admission, study, prescription, and diagnosis is retained within the linkable population defined in row four.
The evaluation cohorts below that row are selected only by task requirements, such as having an ICU stay or an occult condition.
ICU stays are drawn from the MIMIC-IV \texttt{icu} module as the first stay per admission.
Eligibility requires $\min(\text{deathtime},\text{dischtime})-\text{intime}\geq48$ hours, so the in-hospital mortality label postdates every current-stay input.
Laboratory values, medications, and radiographs enter only from the first 48 hours, and longitudinal history includes only admissions strictly before the current admission.

\subsection{MIMIC modality construction}

Each admission connects to diagnoses, prescriptions, procedures, time-bounded laboratory features, and chest-radiograph studies.
Diagnoses and prescriptions are preserved as typed patient-side observations before their standardized concepts are aligned with PrimeKG.
Laboratory measurements remain attached to their admission and observation window, including their values, units, timestamps, and source-table identifiers.

Each chest radiograph is processed with BiomedCLIP \citep{zhang2023biomedclip} to produce scores for named radiographic findings.
Finding-specific thresholds are calibrated against CheXpert reference labels \citep{irvin2019chexpert}.
Findings that pass their thresholds become textual \texttt{imaging-finding} nodes linked to the source radiograph and study.
The ontology aligner then maps these finding nodes to compatible phenotype concepts in PrimeKG.
This image-to-finding-to-concept path allows image content to participate in graph retrieval without discarding the original radiograph or its provenance.

Radiology reports are processed independently.
Report-derived findings become separate observation nodes connected through \texttt{reported-finding} edges.
Consequently, an image-derived finding and a report-derived finding remain distinguishable even when they align to the same biomedical concept.
This separation supports channel ablations and allows retrieved evidence to be traced to either the image model or the clinical report.
Entity-linking routes and coverage are described in App.~\ref{app:linking}.

\FloatBarrier
\section{Entity linking details}
\label{app:linking}
This section provides the terminology routes and mapping-coverage details summarized in \S\ref{sec:alignment}.
Diagnoses are mapped from ICD-9/10 to UMLS CUIs and then to MONDO or HPO using UMLS 2026AA.
Exact source mappings cover 4{,}168 codes; a synonym hop through MRREL adds 3{,}565; a guarded \texttt{isa}/RB parent hop adds 2{,}238; and a blocklisted name match adds 266.
Coverage reaches 80.5\% of disease diagnosis rows (64.2\% of all rows; administrative Z/V/Y codes are excluded from the knowledge join).
The ICD-to-CUI table must be keyed by \emph{(version, code)}, not by code alone.
Twenty-one code strings have different meanings in ICD-9 and ICD-10; for example, E8341 denotes a transport accident in one version and hypermagnesemia in the other.
Prescriptions map to DrugBank accessions through normalized names, UMLS synonyms, and NDC routes.
We first remove 110 formulation and vehicle strings, including diluents and flushes, which account for 19.7\% of raw prescription rows.
The pipeline then maps 91.1\% of the remaining prescription volume (47.9\% of distinct strings), and 92.8\% of the mapped volume has an ATC class.
Coverage figures in this section use bridge v2; the frozen graph and benchmark use bridge v1 (App.~\ref{app:split-audit}).

\FloatBarrier
\section{Evaluation settings, models, and statistical procedures}
\label{app:evaluation}

\textbf{Outcome prediction.}
The primary MIMIC cohort contains 28{,}582 ICU stays from 18{,}396 patients, including 2{,}306 in-hospital deaths.
We retain the first ICU stay per admission and restrict current-stay inputs to the first 48 hours.
Longitudinal history includes only admissions preceding the current admission.
The zero-shot LLM is evaluated on the full eligible cohort.
The trained GraphSAGE consumer uses a fixed 20\% patient-level test split containing 5{,}676 stays, with no patient shared between the training and test sets.
The primary metric is AUROC.
A separate longitudinal ADNI prognosis experiment contains 619 participants and is evaluated across five seeds using balanced accuracy.
The complete GraphSAGE architecture and training procedure are given in App.~\ref{app:gnn}.

\textbf{Biomedical fact-checking.}
We construct 1{,}920 balanced yes/no questions across eight biomedical relation families from a source cohort of 47{,}100 patients and 183{,}950 admissions.
The experimental conditions compare questions without KG evidence against complete PrimeKG relation lists, entity-only controls, shuffled neighborhoods, and corrupted-name controls.
The evaluation unit is one question, and the primary metric is balanced accuracy.
App.~\ref{app:completeness} describes the retrieval-completeness experiments and reports results for all eight families.

\textbf{Patient-specific join questions.}
The MIMIC benchmark contains 3{,}200 balanced items, comprising 400 matched positive--negative pairs in each of four families.
The factorial design compares a conventional patient record (A), added patient modalities (B), added PrimeKG relations (C), and both channels (D).
The primary drug-controlled analysis uses 1{,}895 items whose candidate drug appears in both label classes.
The ADNI paired contraindication benchmark contains 510 development items, 114 validation items from 61 participants, and a 70-item test set from 39 participants.
The one-time test evaluation provides the primary ADNI cross-cohort result, while the validation set is retained for detailed controls and model-sensitivity analyses.
Detailed benchmark construction and leak controls are provided in App.~\ref{app:benchcons}, and the ADNI evaluation is described in App.~\ref{app:adnieval}.

\textbf{Language models and scoring.}
The primary MIMIC experiments use Qwen3.5-35B-A3B with 4-bit GPTQ quantization, served locally with vLLM \citep{kwon2023vllm}.
The primary ADNI test set evaluation uses Qwen3.5-9B, and the same frozen ADNI validation set with Qwen2.5-7B-Instruct was performed as a model-sensitivity analysis.
Separate ADNI retrieval studies use local Qwen3-8B and Qwen3-32B-FP8 models and are reported separately from the ADNI validation set.
For yes/no questions, we normalize the first-token probabilities over \texttt{yes} and \texttt{no}, $p_{\mathrm{yes}}=P(\texttt{yes})/ [P(\texttt{yes})+P(\texttt{no})]$, and use $p_{\mathrm{yes}}$ as the continuous score; balanced accuracy uses the fixed threshold $p_{\mathrm{yes}}>0.5$.

\textbf{Uncertainty and statistical comparisons.}
MIMIC confidence intervals and comparisons use 2{,}000 paired-item bootstrap samples.
The primary ADNI confidence interval treats participants as clusters, while matched-pair and candidate-drug clustering provide sensitivity analyses.
Factorial interactions use the difference-in-differences definition in \S\ref{sec:factorial}.
Positive interaction indicates that one information source becomes more useful when the other is available.

\FloatBarrier
\section{Controlled join-benchmark construction and leak controls}
\label{app:benchcons}

\subsection{Benchmark items and an example}

\begin{figure}[htbp]
\fbox{\begin{minipage}{0.972\textwidth}\footnotesize
\textbf{Question:} Is vancomycin contraindicated for this 83-year-old ICU patient, given her documented condition? Answer \texttt{yes} or \texttt{no}.\\[2pt]
\textbf{Shared record (A--D):} no coded diagnoses; current medications include vancomycin.\\[2pt]
\textcolor{inkmm}{\textbf{Patient evidence (B,D):} first-window creatinine 4.3--4.4 mg/dL and blood urea nitrogen (BUN) 91--108 mg/dL reveal severe renal impairment while the renal state is not named in the prompt.}\\[2pt]
\textcolor{inkkg}{\textbf{PrimeKG evidence (C,D):} vancomycin's complete contraindication list includes kidney disease.}\\[4pt]
\centering
\setlength{\tabcolsep}{7pt}
\begin{tabular}{@{}lccc@{}}
\toprule
Condition & Patient modalities & PrimeKG relation & Logged answer / score \\
\midrule
A & no & no & \texttt{no} / 0.011 \\
B & yes & no & \texttt{no} / 0.026 \\
C & no & yes & \texttt{no} / 0.002 \\
D & yes & yes & \texttt{yes} / 0.651 \\
\bottomrule
\end{tabular}
\end{minipage}}
\caption{Abridged join-task example; only condition D contains both premises needed for the correct answer, \texttt{yes}.
The complete prompt and its same-patient negative are provided in App. \ref{supp:prompts}}
\label{fig:factorial-conditions}
\end{figure}

\textbf{Benchmark design.}
The \emph{contra-labs} and \emph{contra-report} families connect an occult patient condition to a drug contraindication; \emph{drug-effect} connects a report finding to an adverse effect; and \emph{mechanism} connects a drug target to a protein implicated in an occult condition.
Conditions A--D receive identical questions (Figure~\ref{fig:factorial-conditions}).
Each positive patient--drug pair is matched to a negative drug received by the same patient.
Every negative has a non-empty, length-matched KG list and no near-synonym of a visible patient concept, and all main analyses use complete relation lists (App.~\ref{app:benchcons}).

The eligible patient pools contain 2{,}307 patients for \emph{contra-labs}, 456 for \emph{contra-report}, 2{,}143 for \emph{drug-effect}, and 476 for \emph{mechanism}.
We sample 400 pairs from each family.
Three shortcuts were discovered and closed during construction, in order.
(1)~\emph{Presence leak}: drugs labeled safe often had no KG list, making an absent KG block a perfect predictor of ``no.''
We therefore require every negative item to have a non-empty list.
(2)~\emph{Length leak}: positive lists were approximately three times longer because drugs contraindicated in common organ states tend to have long relation lists.
We match negatives on rendered character count, yielding final total-prompt positive-to-negative length ratios of 0.99--1.03.
(3)~\emph{Near-synonym leak}: labels are defined by PrimeKG identifiers, but models read strings, and distinct nodes can have nearly identical names, such as ``edema'' and ``pulmonary edema.''
We reject a negative if its list contains a subset-containment near-synonym of any patient concept visible in the prompt; the residual rate is 0--1\% per family.
Before control~(3), the drug-effect family was degenerate (specificity 0.04).
A pilot with permissive wording (``could \emph{any} finding\ldots'') also saturated the decision threshold.
The released question instead requires an identified match: ``answer yes only if you can identify a specific finding that appears in the adverse-effect list.''
This wording restores a significant interaction on both metrics, although a tendency to answer yes remains at the 0.5 threshold, as reported in the main text.
The stored ICU prompt used a legacy ``first 24 hours'' header although its values came from the prespecified first-48-hour window.
This wording error is identical in all conditions and therefore does not affect their comparison.
The released prompt corrects the header.
% Example items and full prompts for every condition are included in the supplementary material.
Exact prompt templates and serialization rules for all conditions, together with
redacted matched examples are included in App. \ref{supp:prompts}.

\FloatBarrier
\section{Retrieval completeness}
\label{app:completeness}
Across the three set-membership fact-checking families ($n{=}720$), retrieval capped at 12 edges yields balanced accuracy 0.718, with sensitivity 0.483 and specificity 0.953.
This pattern is consistent with the model treating an omitted relation as absent.
Complete neighborhoods raise balanced accuracy to 0.913, with sensitivity 0.911 and specificity 0.914.
For the side-effect family, truncated retrieval (0.642) performs worse than no retrieval (0.725).
PrimeKG lists a median of 43 side effects per drug and as many as 338, so a 12-edge list omits the queried effect for 78\% of positive items.
Warning the model that the evidence may be incomplete recovers $+0.142$ on one family and provides no gain on the others.
The failure therefore requires more complete retrieval rather than a prompt-only correction.
All main-paper KG conditions therefore use complete lists.

\begin{table}[htbp]
\caption{Biomedical fact-checking with complete retrieval ($n{=}1{,}920$ balanced items).
\textbf{A}: question only; \textbf{B}: entity names; \textbf{C}: retrieved KG edges; C$_{\mathrm{shuf}}$: same-length neighborhood from another item; C$_{\mathrm{corr}}$: true edge structure with corrupted names.
$\Delta$ is C$-$A; all $p<0.001$.}
\label{tab:audit}
\centering
\footnotesize
\setlength{\tabcolsep}{4pt}
\begin{tabular}{lcccccc}
\toprule
Family (hops) & A & B & C & C$_{\mathrm{shuf}}$ & C$_{\mathrm{corr}}$ & $\Delta\!\uparrow$ \\
\midrule
Contraindication (1) & 0.537 & 0.608 & \best{0.958} & 0.525 & 0.500 & $+0.420$ \\
Shared drug target (2) & 0.575 & 0.571 & \best{0.975} & 0.504 & 0.500 & $+0.400$ \\
Mechanistic overlap (2) & 0.592 & 0.617 & \best{0.946} & 0.533 & 0.500 & $+0.355$ \\
Drug side effect (1) & 0.721 & 0.742 & \best{0.983} & 0.533 & 0.529 & $+0.262$ \\
Drug class / ATC (1) & 0.667 & 0.658 & \best{0.871} & 0.604 & 0.508 & $+0.203$ \\
Three-hop chain (3) & 0.604 & 0.592 & \best{0.787} & 0.525 & 0.500 & $+0.184$ \\
Comorbidity (1) & 0.808 & 0.767 & \best{0.938} & 0.579 & 0.717 & $+0.129$ \\
Indication (1) & 0.925 & 0.921 & \best{0.988} & 0.696 & 0.500 & $+0.063$ \\
\bottomrule
\end{tabular}
\end{table}

\begin{table}[htbp]
\caption{Robustness of the join interaction (AUROC) across language models and KG renderings.
$^{\dagger}p\geq0.05$; all other entries $p<0.05$.}
\label{tab:multimodel}
\centering
\footnotesize
\setlength{\tabcolsep}{5pt}
\begin{tabular}{lccccc}
\toprule
& \multicolumn{2}{c}{Qwen3.5-35B} & \multicolumn{2}{c}{Qwen3.5-9B} & Llama-3.1-8B \\
\cmidrule(r){2-3}\cmidrule(r){4-5}\cmidrule(l){6-6}
Family & prose & triples & prose & triples & prose \\
\midrule
contra-report & $+0.228$ & $+0.208$ & $+0.138$ & $+0.142$ & $+0.025^{\dagger}$ \\
contra-labs & $+0.121$ & $+0.101$ & $+0.063$ & $+0.010^{\dagger}$ & $+0.046$ \\
drug-effect & $+0.463$ & $+0.471$ & $+0.513$ & $+0.479$ & $+0.524$ \\
mechanism & $+0.118$ & $+0.185$ & $+0.168$ & $+0.110$ & $+0.084$ \\
\midrule
All & \best{$+0.209$} & \best{$+0.220$} & \best{$+0.201$} & \best{$+0.153$} & \best{$+0.092$} \\
\bottomrule
\end{tabular}
\end{table}

\begin{table}[htbp]
\caption{Mechanism checks.
Left: loss in balanced accuracy after dropping one patient channel with the KG on.
Right: the AUROC interaction appears mainly when the state is absent from coded text. $^{*}p<0.05$, n.s. \ not significant.}
\label{tab:mechanism}
\centering
\footnotesize
\setlength{\tabcolsep}{2.8pt}
\begin{tabular}{lrrr@{\hskip 12pt}lrr}
\toprule
\multicolumn{4}{c}{Leave-one-out (KG on)} & \multicolumn{3}{c}{Occultness split} \\
\cmidrule(r){1-4}\cmidrule(l){5-7}
Family & $-$labs & $-$report & $-$image & Subset ($n$) & C & Interaction \\
\midrule
contra-labs & \best{$+0.090^{*}$} & $+0.004$ & $+0.000$ &
Strictly occult (654) & 0.483 & \best{$+0.126^{*}$} \\
contra-report & $-0.017$ & \best{$+0.162^{*}$} & $+0.005$ &
Readable in codes (146) & 0.783 & $+0.040$ n.s. \\
mechanism & $-0.022$ & \best{$+0.087^{*}$} & $-0.026^{*}$ & & & \\
drug-effect & $-0.119^{*}$ & \best{$+0.087^{*}$} & $-0.010$ & & & \\
\bottomrule
\end{tabular}
\end{table}

\FloatBarrier
\section{Behavioral validation of PrimeKG and layer refinement}
\label{app:validation}
For every drug--condition pair we compute the log odds ratio of prescription given the condition over 183{,}950 admissions.
PrimeKG's indication layer is strongly corroborated: the mean log odds ratio is $+1.13$, compared with $+0.08$ for unlabeled pairs ($p{<}0.001$).
By contrast, contraindication pairs are not avoided as a group ($+0.12$), and their mean is indistinguishable from that of unlabeled pairs.
We split the cohort into disjoint patient halves and require avoidance to replicate across both halves.
This procedure identifies 649 behaviorally corroborated contraindications and 1{,}817 relations that observed prescribing reliably contradicts, with replication rates of 97.6\% and 96.3\% against 50\% chance.
Inspection of the latter group identifies clear errors; for example, PrimeKG lists fondaparinux as contraindicated in heparin-induced thrombocytopenia even though it is a standard treatment.
When we repeat the fact-checking audit on both subsets, the model follows corroborated and contradicted edges similarly ($+0.104$ versus $+0.130$; not significantly different).
We additionally intersect these behavioral partitions with the answer-bearing contraindication edges of the controlled join benchmark, retaining each selected positive item together with its same-admission matched negative.
The behaviorally corroborated subset contains 152 items (76 matched pairs) and retains an AUROC interaction of $+0.145$ (95\% CI $[+0.074,+0.224]$, $p<0.001$).
The behaviorally contradicted subset contains 580 items (290 matched pairs) and gives a nearly identical interaction of $+0.150$ (95\% CI $[+0.111,+0.190]$, $p<0.001$).
% The model can therefore follow the supplied relation without determining whether that relation is clinically correct, which motivates validating KG content before deployment.
Together, these analyses show that the model can join patient evidence to a supplied relation regardless of whether observed prescribing behavior corroborates that relation; they do not establish the clinical correctness of the underlying KG edge.
% Thus, the patient--knowledge interaction does not depend on whether the answer-bearing PrimeKG contraindication is corroborated by observed prescribing behavior.

\FloatBarrier
\section{Imaging channel and the information ceiling}
\label{app:imaging}
\label{app:ladder}
For a chest radiograph $x$, the imaging harmonizer computes finding-specific scores $s_k(x)$ and produces
\begin{equation}
Z_{\mathrm{img}}(x)
=
\{z_k:s_k(x)\geq\eta_k\},
\end{equation}
where $z_k$ is a named radiographic finding and $\eta_k$ is its calibrated threshold.
Each retained finding becomes an observation node linked to its source radiograph and is subsequently aligned with a compatible phenotype concept.
We calibrate the vision labels against CheXpert reference labels on 203{,}999 studies.
Findings with AUROC below 0.70 are removed, and finding-specific thresholds are selected on one calibration half.
The default threshold of 0.5 predicted some findings at up to $21.8\times$ their true prevalence.

For the zero-shot LLM, image-derived findings contribute approximately zero
to mortality prediction in the full context (Table~\ref{tab:prediction}).
A separate single-channel LLM comparison explains this result.
Imaging alone reaches AUROC 0.631, compared with 0.586 for demographics, so
the image labels contain outcome signal.
However, on this comparison cohort, the full record reaches 0.700 with or
without imaging, indicating redundancy for the LLM rather than an absence of
information.
The trained GNN shows a different pattern: removing image features reduces
AUROC by 0.008 ($p{=}0.001$).

We next use the radiologist's structured findings as an information ceiling in a representation ladder ($n{=}7{,}645$): finding names, severity, laterality, interval change and devices, and verbatim prose.
For the image channel alone, the added graded structure improves AUROC by $+0.046$.
Interval change contributes $+0.030$, severity contributes $+0.014$, and laterality contributes no gain.
Once laboratory values and history are present in the LLM prompt, however,
no representation rung outperforms the no-imaging condition.
The structured representation also exceeds the radiologist's verbatim prose by $+0.025$.
For this outcome, structured extraction is more useful to the LLM than raw clinical prose, but even the human-derived information ceiling adds no signal beyond the full record.

\FloatBarrier
\section{Run-to-run noise floor}
\label{app:noise}
For patients lacking a chest film, the full-prompt and no-imaging conditions are identical, yet scores differ across serving runs because batch composition changes floating-point reduction order.
Across 19{,}678--21{,}383 identical-prompt pairs, the AUROC gap between runs is $+0.0003$--$+0.0042$, and the median per-item score drift is $2.5\times10^{-3}$.
We therefore treat $\pm0.004$ AUROC as the noise floor; all interaction estimates exceed it by more than an order of magnitude.
Within-run comparisons (all join-task conditions; all ladder conditions) are scored by a single engine instance and do not incur this term.

\FloatBarrier
\section{GNN experiments}
\label{app:gnn}
The trained consumer in Table~\ref{tab:prediction} is a two-layer heterogeneous GraphSAGE network.
It uses \texttt{HeteroConv} with a per-edge-type \texttt{SAGEConv}, sum aggregation, hidden size 32, and dropout 0.5.
Frozen pretrained features are projected to the hidden size: SapBERT name embeddings for patient-side concept nodes, PubMedBERT description embeddings for PrimeKG nodes, Bio\_ClinicalBERT embeddings for matched radiology reports, and BiomedCLIP features for radiograph studies.
The admission's demographics and first-48-hour laboratory features are concatenated with the pooled graph representation before a two-layer prediction head.

Training uses Adam with learning rate $3\times10^{-3}$, weight decay $10^{-3}$, and batch size 512 for at most 100 epochs.
We use validation early stopping with patience 12.
The test set is a fixed 20\% patient-level stratified split, so no patient appears in both training and test sets.
We ensemble predictions from five training seeds.
Channel ablations remove the corresponding node and edge types and their features.

On 30-day readmission, the GraphSAGE model reaches AUROC 0.644 against 0.548 for tabular logistic regression.
On ICU mortality, the full model reaches AUROC 0.804 against 0.778 for the
laboratory-only comparator; on the imaging-present subset, it reaches 0.854
against 0.819.
In the paired ICU ablations, laboratory values have the largest contribution ($+0.085$), followed by imaging ($+0.008$) and prior-visit history ($+0.005$); all three changes are significant.
Radiology reports, medications, and PrimeKG do not produce measurable gains.
In particular, dropping PrimeKG changes AUROC by only $-0.002$ ($p{=}0.16$), matching the null result from the LLM consumer.

\FloatBarrier
\section{ADNI graph construction, connectivity, and static evaluations}
\label{app:adnigraph}

\textbf{Cohort and domains.}
The full cohort contains 1{,}000 participants across 4{,}977 visits: 284 cognitively normal participants, 39 with significant memory concern, 125 with early mild cognitive impairment (MCI), 275 with MCI, 75 with late MCI, and 202 with Alzheimer's disease (AD).
The initial construction covers imaging, medical history, cognition, genetics, and biospecimens.
Field-level audits add medications, neurological examinations, behavioral inventories, vital signs, and laboratory results.
Every patient-local finding retains its participant, visit, source file and field, construction rule, value direction, and missingness status.
The patient-level split for the static tasks is 800/100/100.
Later language-model studies use the separate item banks described below, and we do not pool their results with the static tasks.

\textbf{MRI harmonization.}
FastSurfer \citep{Henschel2020FastSurfer} processes 8{,}990 MRI series in segmentation-only mode.
Of these, 8{,}987 complete processing and 8{,}984 produce the regional-statistics file used downstream.
For region $k$, the normative model is fitted only on cognitively normal baseline scans in the training split:
\begin{equation*}
V_{ik}=\beta_{0k}+\beta_{1k}\operatorname{Age}_i+\beta_{2k}\operatorname{Sex}_i+\beta_{3k}\operatorname{SupraTentorialVol}_i+\varepsilon_{ik},
\qquad
z_{ik}=\frac{V_{ik}-\widehat V_{ik}}{\widehat\sigma_k}.
\label{eq:adni-normative}
\end{equation*}
Supratentorial volume is the head-size covariate because estimated intracranial volume is unavailable from the segmentation-only run.
Each $|z_{ik}|>2$ event becomes an abnormal imaging finding rather than an attribute merged into an ontology node.

% \textbf{Versioned construction.}
% Table~\ref{tab:adni-versions} reports the patient-layer graph files and the separately materialized concept-relation layer.
% From v6 onward, 9{,}264 additional external concept nodes are stored separately.
% The patient-layer node and edge counts therefore should not be added to the concept-edge count as if they shared one denominator.
% Each version also retains a switch that reproduces the preceding patient graph.

% \begin{table}[htbp]
% \caption{ADNI graph versions.
% Nodes and edges describe the patient-layer graph; the last column describes the separately materialized concept-to-concept layer.}
% \label{tab:adni-versions}
% \begin{center}
% \small
% \setlength{\tabcolsep}{5pt}
% \begin{tabular}{lrrrrrr}
% \toprule
% Version & Findings & CUIs & Concepts & Nodes & Edges & Concept edges \\
% \midrule
% v3 & 41{,}270 & 62 & 64 & 47{,}373 & 170{,}057 & 0 \\
% v4 & 41{,}270 & 62 & 96 & 47{,}405 & 213{,}258 & 0 \\
% v5 & 62{,}019 & 191 & 220 & 68{,}407 & 296{,}254 & 0 \\
% v6 & 62{,}019 & 191 & 220 & 68{,}407 & 296{,}254 & 39{,}080 \\
% v7 & 64{,}786 & 213 & 242 & 71{,}218 & 307{,}322 & 39{,}712 \\
% v8 & 64{,}786 & 213 & 264 & 71{,}240 & 334{,}764 & 41{,}246 \\
% \bottomrule
% \end{tabular}
% \end{center}
% \end{table}

\textbf{Final graph composition.}
The phenotype-anchored ADNI graph contains 64{,}786 retained findings, 213 CUIs, 264 mapped concepts, 71{,}240 patient-layer nodes, and 334{,}764 patient-layer edges.
An additional 9{,}264 external concept nodes and 41{,}246 concept-to-concept edges are stored separately.
The patient-layer and concept-layer counts use different denominators and should not be summed.

\textbf{Coverage did not imply connectivity.}
The expanded Desikan--Killiany--Tourville (DKT) to UBERON crosswalk raised imaging linkage coverage from 35.8\% to 88.4\% and total feature-column coverage from 44.5\% to 86.6\%.
These values measure coverage rather than mapping correctness.
Under anatomical alignment alone, the main concept component contained 9{,}650 of 9{,}905 nodes and all nine non-imaging domains, while all 80 imaging concepts remained outside it in 46 fragments.
Of the 182{,}298 edges touching mapped brain regions, 99.8\% were tissue-expression edges from the noisy layer excluded during graph construction.
No retained path therefore connected imaging concepts to disease, drug, or phenotype concepts.

The phenotype-anchored configuration retained the anatomical \texttt{located-in} edges and added direction-aware \texttt{has-imaging-phenotype} edges.
Ninety-two imaging columns mapped to 22 HPO phenotypes and created 27{,}442 phenotype-anchoring edges.
Twenty of these phenotypes entered the main component and gained two- to four-hop paths to cognitive, behavioral, genetic, and medication concepts.
This was a structural repair rather than a mapping-accuracy result: none of the 96 DKT-to-HPO review rows has been adjudicated, and 60 use lower-confidence regional fallbacks.
For the held-out dynamic benchmark, a separate answerable-concept set maps 112 of 126 concepts, including 104 that resolve to exact PrimeKG nodes.
These values also measure coverage rather than manually verified accuracy.

\textbf{Static and graph-native boundaries.}
On three-class diagnosis, the flat condition reaches macro-F1 $0.6177\pm0.026$.
The concept-mapped patient-graph condition reaches 0.6921, but a flat control model given the same added columns reaches $0.7383\pm0.014$.
The additional data, rather than concept mapping, therefore explain the improvement over the original flat condition.
A discontinued cognitive-slope regression is also negative: the flat model reaches Pearson correlation $0.4034\pm0.023$, whereas earlier graph-feature configurations range from 0.244 to 0.274.
The cognition-dropout experiment suggests greater robustness for the graph condition.
However, it lacks a non-graph concept-pooling control that would separate graph routing from feature dilution, so we do not interpret it as evidence of a graph gain.

The pre-anchor graph-native experiment also gives a null result across ten seeds.
The matched no-graph network reaches macro-F1 $0.6087\pm0.0215$, and the four-layer graph-attention network reaches $0.6109\pm0.0529$.
Because the pre-anchor graph contains no cross-modal concept path from imaging to another domain, this experiment cannot distinguish an ineffective consumer from a graph with no useful structure to propagate.
A phenotype-anchored graph-native rerun is therefore required before the phenotype repair can support a message-passing claim.

\FloatBarrier
\section{ADNI paired contraindication benchmark: validation, test evaluation, and evidence controls}
\label{app:adnieval}

\textbf{Construction.}
The candidate-conditioned retriever is deterministic rule-based code.
It normalizes the candidate drug, applies family-specific relation whitelists, and retrieves its complete bounded PrimeKG packet.
It has no access to the answer-bearing medical-history block or the hidden label.
The median, 90th-percentile, and maximum packet lengths are 623, 947, and 1{,}298 tokens.
The paired contraindication benchmark uses cross-patient, same-candidate positive and negative pairs with identical KG packets and at most two items per participant.
A positive item requires a displayed mapped condition and a contraindication edge between the candidate and that condition.
A negative item means only that none of the mapped conditions displayed in the record has such an edge.

The dynamic study contains 510 development items from 265 participants, 114-item validation from 61 participants, and a 70-item test set from 39 participants, scored once with the frozen Qwen3.5-9B pipeline after the validation analysis was fixed.
The validation bank contains 57 matched pairs and 53 candidate drugs; only three candidates occur in more than one pair.
Qwen3.5-9B is served locally with vLLM in bfloat16, and constrained yes/no probabilities provide the continuous score.
The primary confidence interval clusters by participant.
Pair-clustered, candidate-clustered, and equally weighted candidate analyses serve as sensitivity checks.

\textbf{Interaction and controls.}
With Qwen3.5-9B, validation AUROC is 0.5636 for A (neither channel), 0.5534 for B (patient only), 0.5506 for C (PrimeKG only), and 0.8590 for D (both).
The interaction is $+0.3186$ with participant-clustered 95\% CI $[+0.1806,+0.4539]$; pair- and candidate-clustered analyses are also positive and exclude zero.
The aligned condition exceeds shuffled PrimeKG (0.5131), a degree/relation-matched null (0.5260), decisive-support removal (0.6316), and node names without relations (0.5531).
\begin{table}[htbp]
\caption{\textbf{ADNI paired contraindication benchmark across language
models.}
Both models are evaluated on the same 114-item validation bank using the
same item keys, arm definitions, KG packets, and non-KG inputs.
Interactions are $I_M$ with participant-clustered 95\% confidence
intervals.}
\label{tab:adni-model-comparison}
\centering
\footnotesize
\setlength{\tabcolsep}{4pt}
\begin{tabular}{lrrrrr}
\toprule
Model & A & B & C & D & Interaction [95\% CI] \\
\midrule
Qwen2.5-7B-Instruct
& 0.491
& 0.563
& 0.479
& 0.679
& $+0.1285$ $[+0.0185,+0.2372]$ \\

Qwen3.5-9B
& 0.5636
& 0.5534
& 0.5506
& 0.8590
& $+0.3186$ $[+0.1806,+0.4539]$ \\
\bottomrule
\end{tabular}
\end{table}

\begin{table}[htbp]
\caption{\textbf{ADNI paired contraindication test-set results} ($n=70$ items; 35 positive--negative pairs from 39 participants).
AUROC intervals are participant-clustered 95\% confidence intervals.
The factorial interaction is $I_M =+0.2988$
(95\% CI $[+0.0650,+0.5319]$; predeclared $p=0.0135$).}
\label{tab:adni-confirmatory}
\centering
\footnotesize
\setlength{\tabcolsep}{4pt}
\begin{tabular}{lcc}
\toprule
Condition & AUROC & Participant-clustered 95\% CI \\
\midrule
A: base                         & 0.5584 & $[0.4050,0.7151]$ \\
B: patient only                 & 0.5706 & $[0.4016,0.7305]$ \\
C: knowledge only               & 0.5494 & $[0.3946,0.6988]$ \\
D: patient + dynamic KG         & \best{0.8604} & $[0.7433,0.9557]$ \\
Shuffled KG                     & 0.5314 & $[0.3949,0.6741]$ \\
Degree/relation null            & 0.5478 & $[0.3979,0.6933]$ \\
Decisive support removed        & 0.7143 & $[0.5676,0.8448]$ \\
Node names only                 & 0.5584 & $[0.3956,0.7040]$ \\
\bottomrule
\end{tabular}
\end{table}

\begin{table}[htbp]
\caption{\textbf{Destructive-control contrasts on the ADNI test set.}
Differences are D minus each control; Holm adjustment covers the four
prespecified destructive-control comparisons.}
\label{tab:adni-confirmatory-controls}
\centering
\footnotesize
\begin{tabular}{lccc}
\toprule
Contrast & $\Delta$AUROC & 95\% CI & Holm $p$ \\
\midrule
D $-$ shuffled KG
& $+0.3290$ & $[+0.1484,+0.4887]$ & 0.0339 \\

D $-$ degree/relation null
& $+0.3127$ & $[+0.1742,+0.4445]$ & 0.0290 \\

D $-$ decisive support removed
& $+0.1461$ & $[+0.0789,+0.2192]$ & $0.0002$ \\

D $-$ node names only
& $+0.3020$ & $[+0.1637,+0.4534]$ & 0.0200 \\
\bottomrule
\end{tabular}
\end{table}

\textbf{Modality localization.}
On validation, the full patient representation reaches AUROC 0.5546 without KG and 0.8529 with the correct KG packet.
Medical history alone with the KG reaches 0.8313, while removing all patient information yields 0.5003 without KG and 0.5046 with KG.
The medical-history enablement estimate is $+0.2936$ (95\% CI $[+0.1618,+0.4237]$) and is the only tested patient block whose magnitude exceeds the prespecified 0.137 minimum detectable interaction.
Independent repeats change no endpoint by more than 0.0022, below the frozen 0.010 reproducibility threshold.

\begin{table}[htbp]
\caption{Exploratory external-method comparison on the ADNI paired contraindication benchmark validation bank ($n=114$). Methods use their method-specific evidence construction; the direct PrimeKG lookup is the primary no-model comparator.
$^{a}$\graphname{} condition D is the previously frozen Qwen3.5-9B patient-plus-KG result on the same validation bank; its complete candidate KG packet differs from the evidence representations used by MindMap and GraphCare, so the row is shown for within-bank context rather than as a controlled architecture-only comparison.}
\label{tab:adni-external}
\centering
\footnotesize
\begin{tabular}{lr}
\toprule
Method & AUROC \\
\midrule
\graphname{} (condition D)$^{a}$ & 0.8590 \\
\midrule
MindMap v2 & 0.9674 \\
Direct PrimeKG lookup & 0.9561 \\
GraphCare GIN adapted & 0.9437 \\
GraphCare BAT adapted & 0.8122 \\
Candidate-only prior & 0.5000 \\
\bottomrule
\end{tabular}
\end{table}
\textbf{Longitudinal multimodal prognosis.}
A separate five-seed study asks whether memory declines faster over roughly four years than the patient's baseline predicts.
Among 619 participants, clinical history and examination alone reach balanced accuracy 0.595, while adding biomarkers and imaging reaches 0.652, a $+0.057$ change with the same sign in all five seeds.
The graph-augmented counterparts reach 0.577 and 0.622.
Both graph increments change sign across seeds and remain smaller than the variation across seeds.
We therefore interpret only the multimodal increment as a consistent effect.
The label is based on follow-up rather than baseline diagnosis; App. source files record the rejected label variants and their baseline and follow-up correlations.

\textbf{Development-stage laboratory/vital-sign join.}
An earlier pooled family contains 194 matched items from 86 participants.
Laboratory values or vital signs indicate a condition absent from the coded record, and the question asks whether that condition is contraindicated with a current drug.
Balanced accuracy is 0.505, 0.541, 0.500, and 0.701 for A--D; the interaction is $+0.165$ with patient-clustered 95\% CI $[+0.101,+0.225]$.
However, the family has no protected test split, and its design choices were made while examining the pooled bank.
We therefore report it as development evidence; the ADNI test set above provides the primary cross-cohort result.
Here ``occult'' means absent from an ADNI enrollment checkbox, not a condition missed by a clinician.

\textbf{ADNI explanation and provenance audit.}
We conducted a separate post-hoc explanation audit on the 114-item
Qwen3.5 validation bank. Condition D produced valid structured output for
all 114 items. Across 85 attempted KG citations, every cited relation was
present in the supplied packet; 53 cited the decisive answer-bearing edge,
32 cited another valid in-packet edge, and no unsupported KG citation was
observed. Relative to the 57 positive items, decisive-edge citation recall
was 53/57 (0.930).

Removing the decisive supporting edge did not induce fabrication of the
missing relation. All 114 support-removed outputs remained structurally
valid; among the 57 positive items, the model cited another available in-packet relation in 48 cases and abstained in 9, and it never cited the removed identifier. The positive-item citation-attempt rate decreased from 55/57 to 48/57 ($\Delta=-0.123$,
95\% CI $[-0.204,-0.039]$).
Removing only provenance headers leaves predictive performance essentially unchanged: AUROC is 0.8587 with the original D prompt and 0.8480 without provenance headers, for a paired difference of $-0.0108$ (95\% CI $[-0.0579,+0.0322]$).
Pair- and candidate-clustered intervals likewise include zero.
Thus, provenance labels improve traceability without being responsible for the observed predictive gain.

\FloatBarrier
\section{ADNI retrieval-regime stress test}
\label{app:adniretrieval}

\textbf{Design.}
The hop ladder contains five balanced families with 100 items each and manipulates only shortest-path length.
The three one-hop families are the individual legs of the longer drug--protein--disease--disease chain.
The closed-book condition supplies no graph.
Static context serializes a relation-guided packet, while the query loop lets the model issue one \texttt{neighbors}, \texttt{find}, or \texttt{answer} action per round.
Hop exclusivity is checked during item construction, and positives and negatives are matched on evidence length.
The experiment uses full PrimeKG because the cohort-specific phenotype-anchored graph is materialized only one hop from cohort seeds.
It therefore evaluates graph access rather than the sufficiency of the ADNI cohort-specific graph.

\begin{table}[htbp]
\caption{Balanced accuracy for three graph-access regimes in the ADNI-derived hop ladder. Gold-path presence is measured in the static packet.}
\label{tab:adni-hop}
\begin{center}
\small
\setlength{\tabcolsep}{5pt}
\begin{tabular}{lrrrrr}
\toprule
Relation family & Hops & Closed book & Static context & Query loop & Gold path \\
\midrule
Drug--protein & 1 & 0.760 & 0.370 & 0.920 & 100\% \\
Protein--disease & 1 & 0.570 & 0.400 & 0.970 & 100\% \\
Disease--disease & 1 & 0.820 & 0.290 & 0.980 & 100\% \\
Drug--protein--disease & 2 & 0.450 & 0.430 & 0.670 & 100\% \\
Drug--protein--disease--disease & 3 & 0.470 & 0.450 & 0.590 & 90\% \\
\bottomrule
\end{tabular}
\end{center}
\end{table}

\textbf{The regimes fail differently.}
Closed-book performance falls to chance at two and three hops, showing that the model does not reliably supply the missing relations from its parameters.
Static context falls below closed book in every family even though the gold path is usually present.
At two hops, it answers yes on 82\% of items and has specificity 0.080.
The query loop is strong at one hop but reaches only 0.670 and 0.590 at two and three hops.
At two hops, its sensitivity is 0.340 because failed retrieval is often interpreted as evidence that a relation is absent.
Complete retrieval is therefore necessary but not sufficient, and the two graph-enabled regimes fail in opposite directions.

\textbf{Qualitative traces.}
On a one-hop enzyme question, the larger Qwen3 model independently queries the enzyme neighbors of celecoxib and warfarin, intersects the two lists, and correctly identifies CYP2C8, CYP2C9, and CYP3A4.
On a whole-regimen question, the smaller model repeats the same truncated disease-neighborhood query for all 15 rounds, never queries from the drug side, and returns the wrong answer.
A paired two-hop example shows a complementary static-context failure.
The packet contains the positive path from lutein through ALB to gastric neoplasm, but the model also answers yes to a negative naproxen--plasma-cell-myeloma item with no such path.
These examples show why a retrieved answer should be checked against an explicit path rather than treated as grounded simply because graph text was present.

\textbf{Result boundaries.}
The hop ladder is a knowledge-access stress test, not a clinical outcome task, and it uses full PrimeKG rather than the materialized ADNI graph.
The reported language-model runs also require an archived repeat-run noise estimate.
In addition, the 32-token constrained-decoding configuration produced whitespace-only truncations before the answer field in some earlier studies.
These issues must be resolved before interpreting small differences between retrieval policies.

\FloatBarrier
\section{Additional frozen evaluation results}
\label{app:new-results}

\subsection{Audit boundary for the retrospective MIMIC split}
\label{app:split-audit}

The original 3{,}200-item join benchmark was constructed without a development--test split, and shortcut controls were tuned while inspecting the complete item bank.
We subsequently created a patient-disjoint retrospective split (development: 802 items from 360 patients; test: 2{,}398 items from 1{,}079 patients), selected policies and thresholds on development, froze all configurations, and scored the test split once.
The held-out results therefore test model and retrieval-policy generalization within a previously designed benchmark; they are not an independent validation of benchmark construction.
The patient-side block used by these policies is admission-scoped rather than question-specific, while the PrimeKG block is candidate- and relation-conditioned.
The frozen graph and benchmark use the bridge-v1 alignment, whereas separately quoted MIMIC mapping-coverage audits use bridge v2; coverage percentages from bridge v2 should not be treated as properties of the frozen benchmark prompts.
Finally, in 4 of 1{,}600 positive items, the answer-bearing report finding falls beyond the frozen prompt's 20-name report cap.
These prompts were not changed after freezing, so those cases remain in the evaluation as unavailable patient evidence.

\subsection{Held-out MIMIC retrieval policies}
\label{app:mimic-retrieval}

\begin{table}[htbp]
\caption{\textbf{Retrieval-policy results on the retrospective MIMIC test split}
($n{=}2{,}398$ items; 1{,}079 patients).
R0--R5 are the frozen retrieval policies; MindMap-PrimeKG is included as an
external comparator evaluated on the same test items.
Input is the median number of model-input tokens.
Path recall is the fraction of the 1{,}199 positive items for which the gold
patient evidence is visible under the frozen patient-side caps and every edge
of at least one gold answer path is retrieved.
$\Delta$ and Holm-adjusted $p$ values compare each policy with R4.
The final column gives retrieval/use failures among the 1{,}199 positive items.}
\label{tab:retrieval-test}
\centering
\scriptsize
\setlength{\tabcolsep}{2.5pt}
\resizebox{\textwidth}{!}{%
\begin{tabular}{llrrrrrc}
\toprule
Policy &
Configuration &
Input &
Path recall &
AUROC [95\% CI] &
$\Delta$ &
Holm $p$ &
Failures (retrieval / use) \\
\midrule

R5 &
Gold-path oracle &
660 &
0.997 &
0.917 $[0.906,0.927]$ &
$+0.186$ &
$<0.001$ &
4 / 251 \\

R4 &
\graphname{} &
1{,}206 &
0.997 &
\best{0.731 $[0.710,0.750]$} &
ref. &
--- &
4 / 442 \\

R2 &
SapBERT, KG-only, 8K &
8{,}157 &
0.746 &
0.712 $[0.691,0.733]$ &
$-0.019$ &
0.010 &
305 / 263 \\

MindMap &
evidence + yes-token &
4{,}448 &
0.726 &
0.686 $[0.665,0.708]$ &
$-0.044$ &
$<0.001$ &
329 / 401 \\

R2 &
SapBERT, multi-query, 8K &
8{,}157 &
0.649 &
0.669 $[0.647,0.691]$ &
$-0.062$ &
$<0.001$ &
421 / 216 \\

R3 &
Typed one-hop, 32K &
32{,}353 &
0.747 &
0.655 $[0.633,0.677]$ &
$-0.076$ &
$<0.001$ &
304 / 373 \\

R1 &
Full admissible, 32K &
32{,}420 &
0.982 &
0.650 $[0.629,0.672]$ &
$-0.081$ &
$<0.001$ &
22 / 463 \\

R1 &
Full admissible, 8K &
8{,}156 &
0.736 &
0.648 $[0.626,0.670]$ &
$-0.082$ &
$<0.001$ &
316 / 312 \\

R1 &
Full admissible, 4K &
4{,}080 &
0.549 &
0.646 $[0.623,0.667]$ &
$-0.085$ &
$<0.001$ &
541 / 280 \\

R2 &
SapBERT, KG-only, matched &
1{,}189 &
0.257 &
0.638 $[0.616,0.660]$ &
$-0.092$ &
$<0.001$ &
891 / 69 \\

R3 &
Typed one-hop, 8K &
8{,}154 &
0.735 &
0.634 $[0.612,0.656]$ &
$-0.097$ &
$<0.001$ &
318 / 343 \\

MindMap &
graph-of-thought generation &
4{,}409 &
0.726 &
0.631 $[0.612,0.651]$ &
$-0.100$ &
$<0.001$ &
329 / 288 \\

R3 &
Typed one-hop, 4K &
4{,}080 &
0.629 &
0.626 $[0.604,0.648]$ &
$-0.104$ &
$<0.001$ &
445 / 280 \\

R3 &
Typed one-hop, 2K &
2{,}032 &
0.194 &
0.574 $[0.551,0.597]$ &
$-0.157$ &
$<0.001$ &
966 / 38 \\

R1 &
Full admissible, matched &
1{,}166 &
0.001 &
0.565 $[0.542,0.589]$ &
$-0.166$ &
$<0.001$ &
1{,}198 / 0 \\

R1 &
Full admissible, 2K &
2{,}032 &
0.155 &
0.553 $[0.530,0.577]$ &
$-0.177$ &
$<0.001$ &
1{,}013 / 103 \\

R0 &
Closed book &
74 &
0.000 &
0.548 $[0.526,0.572]$ &
$-0.182$ &
$<0.001$ &
1{,}199 / 0 \\

R3 &
Typed one-hop, matched &
1{,}191 &
0.018 &
0.526 $[0.504,0.549]$ &
$-0.204$ &
$<0.001$ &
1{,}177 / 4 \\

\bottomrule
\end{tabular}}
\end{table}

On the drug-balanced sensitivity subset ($n{=}1{,}418$), R4 reaches AUROC 0.695 (95\% CI $[0.666,0.722]$) with 1{,}217 tokens, while the strongest R2 policy reaches 0.693 ($[0.665,0.721]$) with 8{,}161 tokens ($\Delta=-0.002$, Holm-adjusted $p{=}0.886$).
The oracle reaches 0.920 ($[0.904,0.934]$).

%\begin{wrapfigure}[31]{r}{0.68\textwidth}
\begin{figure}
    \centering
    \includegraphics[width=0.6\textwidth]{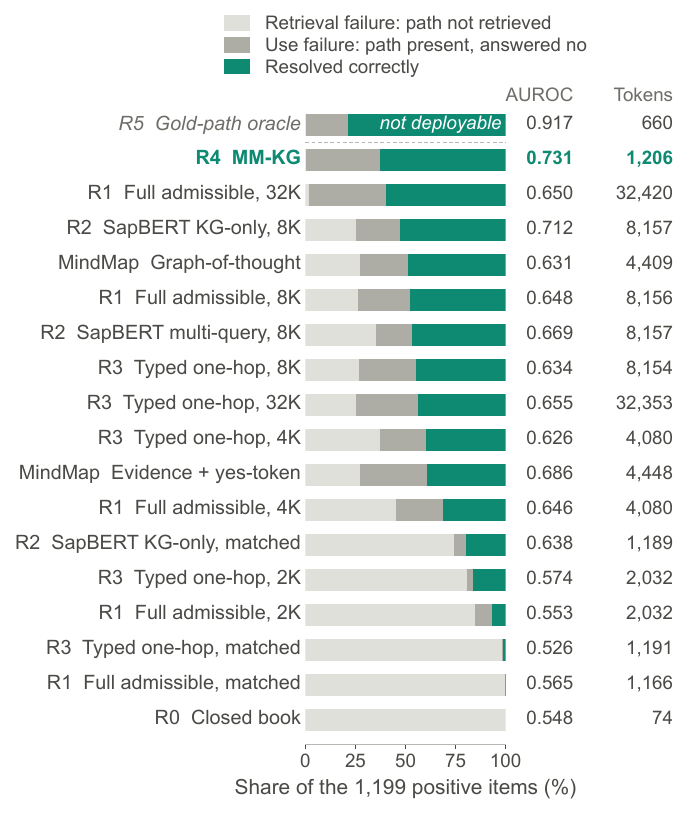}
    \caption{\textbf{All sixteen frozen retrieval policies and two MindMap comparator configurations, ordered by the proportion of positive items resolved.} Each bar decomposes the 1,199 positive items of the frozen MIMIC test split into the answer path never being retrieved, the path being present while the model still answers negatively, and the remainder, resolved correctly, ordering rows by the share of the 1,199 positive items each policy resolves correctly rather than by AUROC. The two orderings are not equivalent: the 32K full-admissible policy resolves more positive items than the strongest vector policy despite a lower AUROC, because this decomposition counts only positive items at the decision threshold whereas AUROC is threshold-free and also reflects performance on the negatives.}
    \label{fig:failure-decomposition-full}
\end{figure}

\FloatBarrier
\subsection{Held-out candidate ranking, controls, and explanation diagnostics}
\label{app:mimic-ranking}

\textbf{An earlier five-packet rendering provides a separate audit of explanation traceability.}
With provenance present, 98.2\% of outputs are valid JSON and 79.5\% of cited edges exist in PrimeKG, but only 16.5\% reproduce the exact answer path; evidence precision and recall are 0.312 and 0.447.
Removing provenance headers unexpectedly raises Hits@1 to 0.492 ($\Delta=+0.030$, Holm-adjusted $p{=}0.016$), while evidence precision and recall become zero because source tags are unavailable.
Deleting the gold edge drives condition-identification accuracy to 0.007 and exact-path recovery to 0.001.
Provenance therefore makes supplied evidence inspectable, whereas correct semantic alignment is what supports the decision; neither property establishes that the underlying KG relation is clinically valid.

\begin{table}[htbp]
\caption{Frozen candidate-ranking results on MIMIC ($n{=}1{,}244$ independent patients; chance Hits@1 0.200).
MM-KG uses the per-candidate query-conditioned R4q formulation.
Marginal Hits@1 intervals and paired $\Delta$ intervals use
2{,}000 bootstrap resamples of the 1{,}244 independent patients;
Holm-adjusted $p$ values compare each row with condition D.
The identity-hard subset ($n{=}598$) contains items for which a candidate-only prior based on training-bank drug frequencies ranks the wrong drug. 
Structured-explanation diagnostics were evaluated separately under the earlier five-packet rendering and are not mixed with the R4q results here.}
\label{tab:ranking-test}
\centering
\scriptsize
\setlength{\tabcolsep}{2.5pt}
\resizebox{\textwidth}{!}{%
\begin{tabular}{llrrrrr}
\toprule
Group &
Condition / policy &
Hits@1 [95\% CI] &
$\Delta$ vs. D [95\% CI] &
Holm $p$ &
MRR &
Identity-hard \\
\midrule

Factorial &
A: neither channel &
0.293 $[0.269,0.319]$ &
$-0.281$ $[-0.318,-0.245]$ &
$<0.001$ &
0.539 &
0.268 \\

Factorial &
B: patient only &
0.304 $[0.277,0.330]$ &
$-0.271$ $[-0.307,-0.236]$ &
$<0.001$ &
0.555 &
0.276 \\

Factorial &
C: PrimeKG only &
0.278 $[0.254,0.304]$ &
$-0.297$ $[-0.328,-0.265]$ &
$<0.001$ &
0.523 &
0.296 \\

Factorial &
D: patient + PrimeKG (R4q) &
\best{0.575 $[0.548,0.603]$} &
ref. &
--- &
0.738 &
\best{0.582} \\

\midrule

Disruption &
Gold edge removed &
0.299 $[0.273,0.323]$ &
$-0.276$ $[-0.305,-0.250]$ &
$<0.001$ &
0.540 &
0.279 \\

Disruption &
Matched wrong packet &
0.237 $[0.213,0.261]$ &
$-0.338$ $[-0.371,-0.303]$ &
$<0.001$ &
0.484 &
0.253 \\

Disruption &
Shuffled packets &
0.185 $[0.162,0.206]$ &
$-0.390$ $[-0.425,-0.355]$ &
$<0.001$ &
0.451 &
0.176 \\

\midrule

Patient channel &
No laboratories &
0.427 $[0.400,0.455]$ &
$-0.148$ $[-0.177,-0.120]$ &
$<0.001$ &
0.627 &
0.415 \\

Patient channel &
No report &
0.452 $[0.424,0.479]$ &
$-0.123$ $[-0.150,-0.097]$ &
$<0.001$ &
0.652 &
0.448 \\

Patient channel &
No imaging &
0.590 $[0.562,0.616]$ &
$+0.015$ $[-0.003,+0.034]$ &
0.21 &
0.747 &
0.599 \\

Patient channel &
No codes &
0.596 $[0.569,0.623]$ &
$+0.021$ $[-0.001,+0.042]$ &
0.18 &
0.749 &
0.600 \\

Patient channel &
No medications &
0.611 $[0.583,0.637]$ &
$+0.036$ $[+0.011,+0.060]$ &
0.015 &
0.761 &
0.625 \\

Audit channel &
No provenance headers &
0.602 $[0.575,0.628]$ &
$+0.027$ $[+0.006,+0.048]$ &
0.044 &
0.754 &
0.602 \\

\midrule

External &
MindMap-PrimeKG (score) &
0.444 $[0.416,0.472]$ &
$-0.131$ $[-0.163,-0.100]$ &
$<0.001$ &
0.646 &
0.448 \\

External &
GraphCare-BAT adapted &
0.609 $[0.582,0.636]$ &
$+0.034$ $[-0.006,+0.072]$ &
0.21 &
0.764 &
0.375 \\

External control &
GraphCare-BAT, patient removed &
0.645 $[0.618,0.671]$ &
$+0.070$ $[+0.033,+0.107]$ &
$<0.001$ &
0.787 &
0.381 \\

\bottomrule
\end{tabular}}
\end{table}

For transparency, the earlier five-packet rendering of condition D reached Hits@1 0.461 (95\% CI $[0.433,0.488]$; MRR 0.665).
The per-candidate R4q formulation was selected on development before its test score was examined and increased test Hits@1 by +0.114 (95\% CI $[+0.084,+0.146]$, $p<0.001$).

\begin{table}[htbp]
\caption{Strict-matching d8 sensitivity ($n{=}380$; chance Hits@1 0.200).
The ordering is preserved, but the D--A comparison is not conclusive at this sample size ($p{=}0.060$).}
\label{tab:ranking-d8}
\centering
\footnotesize
\setlength{\tabcolsep}{7pt}
\begin{tabular}{lrrrr}
\toprule
Condition & Hits@1 [95\% CI] & Hits@3 & MRR & Candidate AUROC \\
\midrule
A: neither channel & 0.342 $[0.297,0.390]$ & 0.726 & 0.567 & 0.581 \\
B: patient only & 0.250 $[0.208,0.292]$ & 0.766 & 0.520 & 0.606 \\
C: PrimeKG only & 0.274 $[0.226,0.318]$ & 0.747 & 0.532 & 0.564 \\
D: patient + PrimeKG & \best{0.408 $[0.361,0.458]$} & 0.818 & 0.626 & 0.672 \\
\bottomrule
\end{tabular}
\end{table}

\clearpage
\subsection{ADNI validation table and targeted prediction boundary}
\label{app:adni-extra}

\begin{table}[htbp]
\centering
\caption{The join interaction across every subset, model, rendering and cohort tested. All entries are AUROC interactions $I_M$ as defined in \S\ref{sec:factorial}; positive means the two evidence channels are superadditive. $^{***}p<0.001$; $^{*}p<0.05$; n.s.\ not significant.
$^{a}$\S\ref{sec:results-composite} reports the interaction as positive and significant for every family on both metrics without printing per-family $p$ values. $^{b}$Per the Table~\ref{tab:multimodel} caption, every entry without a dagger satisfies $p<0.05$.
$^{c}$Excluding the drug-effect family. 
$^{d}$Behavioral subsets retain a positive item when its answer-bearing
contraindication edge belongs to the independently defined prescribing-behavior
partition and retain that item's same-admission matched negative.
The corroborated subset contains 76 positive--negative pairs; the contradicted
subset contains 290 pairs.
% Confidence intervals are printed only where the paper reports them: the MIMIC paired-item bootstrap intervals exist in the frozen result files but are not in the paper, and are requested in \texttt{OPEN\_QUESTIONS.md}.
}
\label{tab:interaction-robustness}
\footnotesize
\setlength{\tabcolsep}{4pt}
\begin{tabular}{l r l l l}
\toprule
Analysis & $n$ & Interaction & 95\% CI & Source \\
\midrule
\multicolumn{5}{l}{\emph{Primary}}\\
\quad Pooled, all items
    & 3,200
    & $+0.209^{***}$
    & $[+0.191, +0.227]$
    & \S\ref{sec:results-composite} \\

\quad Drug-controlled
    & 1,895
    & $\mathbf{+0.194}^{***}$
    & $[+0.170, +0.217]$
    & \S\ref{sec:results-composite} \\

\quad Leave-one-family-out$^{c}$
    & 2,400
    & $+0.179^{***}$
    & $[+0.160, +0.197]$
    & \S\ref{sec:results-composite} \\

\quad Behaviorally corroborated contraindications$^{d}$
    & 152
    & $+0.145^{***}$
    & $[+0.074, +0.224]$
    & App.~\ref{app:validation} \\

\quad Behaviorally contradicted contraindications$^{d}$
    & 580
    & $+0.150^{***}$
    & $[+0.111, +0.190]$
    & App.~\ref{app:validation} \\

\quad contra-report
    & 800
    & $+0.228^{***}$
    & $[+0.192, +0.265]$
    & \S\ref{sec:results-composite} \\

\quad contra-labs
    & 800
    & $+0.121^{***}$
    & $[+0.090, +0.151]$
    & \S\ref{sec:results-composite} \\

\quad mechanism
    & 800
    & $+0.118^{***}$
    & $[+0.082, +0.154]$
    & \S\ref{sec:results-composite} \\

\quad drug-effect
    & 800
    & $+0.463^{***}$
    & $[+0.423, +0.502]$
    & \S\ref{sec:results-composite} \\

\quad Strictly occult
    & 654
    & $+0.126^{***}$
    & $[+0.088, +0.165]$
    & Table~\ref{tab:mechanism} \\

\quad Readable in codes
    & 146
    & $+0.040$ n.s.
    & $[-0.009, +0.092]$
    & Table~\ref{tab:mechanism} \\

\quad Qwen3.5-35B, triples
    & 3,200
    & $+0.220^{***}$
    & $[+0.199, +0.239]$
    & Table~\ref{tab:multimodel} \\

\quad Qwen3.5-9B, prose
    & 3,200
    & $+0.201^{***}$
    & $[+0.182, +0.219]$
    & Table~\ref{tab:multimodel} \\

\quad Qwen3.5-9B, triples
    & 3,200
    & $+0.153^{***}$
    & $[+0.134, +0.173]$
    & Table~\ref{tab:multimodel} \\

\quad Llama-3.1-8B, prose
    & 3,200
    & $+0.092^{***}$
    & $[+0.074, +0.111]$
    & Table~\ref{tab:multimodel} \\

\addlinespace[2pt]
\multicolumn{5}{l}{\emph{Cross-cohort}}\\

\quad ADNI test set (Qwen3.5-9B)
    & 70
    & $\mathbf{+0.2988}^{*}$
    & $[+0.0650,+0.5319]$
    & Table~\ref{tab:adni-confirmatory} \\
    
\quad ADNI test set (Qwen2.5-7B)
    & 70
    & $+0.107$ n.s.
    & $[-0.070,+0.289]$
    & Table~\ref{tab:composite} \\
    
\quad ADNI validation set (Qwen3.5-9B)
    & 114
    & $\mathbf{+0.3186}^{***}$
    & $[+0.1806, +0.4539]$
    & Table~\ref{tab:adni} \\
    
\quad ADNI validation set (Qwen2.5-7B)
    & 114
    & $+0.1285^{*}$
    & $[+0.0185,+0.2372]$
    & Table~\ref{tab:adni-model-comparison} \\

% \addlinespace[2pt]
% \multicolumn{5}{l}{\emph{Pending experiments}}\\

% \quad ADNI confirmatory bank
%     & 70
%     & \textsc{todo}
%     & ---
%     & App.~\ref{app:adnieval} (unopened) \\
\bottomrule
\end{tabular}
\end{table}

\begin{table}[htbp]
\caption{ADNI validation set ($n{=}114$ items; 57 matched pairs from 61 participants).
A--D form the factorial design; remaining rows disrupt the evidence.
Rows are point estimates; the participant-clustered interaction is $+0.3186$ (95\% CI $[+0.1806,+0.4539]$).}
\label{tab:adni}
\begin{center}
\footnotesize
\setlength{\tabcolsep}{6pt}
\begin{tabular}{lcl}
\toprule
Condition & AUROC$\uparrow$ & Evidence available or disrupted \\
\midrule
A: base & 0.5636 & Neither patient nor KG \\
B: patient only & 0.5534 & Patient record, no KG relation \\
C: PrimeKG only & 0.5506 & KG relation, no patient condition \\
D: patient + PrimeKG & \best{0.8590} & Both aligned channels \\
\midrule
Shuffled PrimeKG & 0.5131 & Wrong candidate packet \\
Degree/relation-matched null & 0.5260 & Nuisance structure without true content \\
Decisive support removed & 0.6316 & Correct packet without answer-bearing edge \\
Node names only & 0.5531 & Entities without relations \\
\bottomrule
\end{tabular}
\end{center}
\end{table}

\begin{table}[htbp]
\caption{Targeted APOE-relation augmentation does not improve ADNI biomarker prediction.
Intervals are 95\% confidence intervals for the AUROC change.}
\label{tab:adni-apoe}
\centering
\footnotesize
\setlength{\tabcolsep}{7pt}
\begin{tabular}{lrrr}
\toprule
Endpoint & Patient representation & $+$ targeted KG & $\Delta$AUROC [95\% CI] \\
\midrule
Amyloid status & 0.777 & 0.773 & $-0.004$ $[-0.027,+0.018]$ \\
Tau status & 0.708 & 0.687 & $-0.021$ $[-0.058,+0.017]$ \\
\bottomrule
\end{tabular}
\end{table}

\FloatBarrier
\section{ADNI data use and acknowledgement}
Data used in preparation of this article were obtained from the Alzheimer's Disease Neuroimaging Initiative (ADNI) database (adni.loni.usc.edu).
As such, the investigators within the ADNI contributed to the design and implementation of ADNI and/or provided data but did not participate in analysis or writing of this report. 
A complete listing of ADNI investigators is available through the official ADNI acknowledgement list.

\paragraph{ADNI acknowledgement}
Data collection and sharing for the Alzheimer's Disease Neuroimaging Initiative (ADNI) is funded by the National
Institute on Aging (National Institutes of Health Grant U19 AG024904). The grantee organization is the Northern
California Institute for Research and Education. In the past, ADNI has also received funding from the National
Institute of Biomedical Imaging and Bioengineering, the Canadian Institutes of Health Research, and private
sector contributions through the Foundation for the National Institutes of Health (FNIH) including generous
contributions from the following: AbbVie, Alzheimer’s Association; Alzheimer’s Drug Discovery Foundation;
Araclon Biotech; BioClinica, Inc.; Biogen; Bristol-Myers Squibb Company; CereSpir, Inc.; Cogstate; Eisai Inc.;
Elan Pharmaceuticals, Inc.; Eli Lilly and Company; EuroImmun; F. Hoffmann-La Roche Ltd and its affiliated
company Genentech, Inc.; Fujirebio; GE Healthcare; IXICO Ltd.; Janssen Alzheimer Immunotherapy Research \&
Development, LLC.; Johnson \& Johnson Pharmaceutical Research \& Development LLC.; Lumosity; Lundbeck;
Merck \& Co., Inc.; Meso Scale Diagnostics, LLC.; NeuroRx Research; Neurotrack Technologies; Novartis
Pharmaceuticals Corporation; Pfizer Inc.; Piramal Imaging; Servier; Takeda Pharmaceutical Company; and
Transition Therapeutics.

\FloatBarrier
% ------------------------------------------------------------------
% Supplementary Section S1: benchmark prompts and worked examples
% Compact companion to App. E (tab:benchcons) and Figure fig:factorial-conditions.
% Real MIMIC prompts contain patient-level values and are not redistributed;
% this section gives the exact templates, redacted matched examples
% ------------------------------------------------------------------
\section{Benchmark prompts and worked examples}
\label{supp:prompts}

\subsection{Common prompt template}
\label{supp:template}

Every item of the controlled join benchmark is scored with one chat prompt.
The system message is fixed across items and conditions:
\begin{PromptVerbatim}
You are a clinical pharmacology expert reviewing an intensive-care medication list.
Answer the question with a single word: yes or no.
\end{PromptVerbatim}
The user message is the concatenation of the blocks admitted by the condition
(App. \ref{supp:blocks}), separated by blank lines, followed by the question.
The chat template is applied with the generation prompt appended and, for
Qwen3.5, thinking disabled.
Decoding is deterministic (temperature 0, one output token, top-20 log-probabilities);
the item score is $P(\texttt{yes})/(P(\texttt{yes})+P(\texttt{no}))$ over the first
output token, and the logged answer is \texttt{yes} when that score exceeds 0.5.

\subsection{Condition A--D block definitions}
\label{supp:blocks}

Each item stores seven text blocks.
The four conditions differ only in which blocks are included; the question and every
included block are identical across conditions.

\begin{table}[t]
\captionsetup{justification=raggedright,singlelinecheck=false}
\caption{Prompt blocks included in factorial conditions A--D.}
\label{tab:supp-prompt-blocks}

\footnotesize
\setlength{\tabcolsep}{4pt}
\renewcommand{\arraystretch}{1.08}

\begin{tabularx}{\linewidth}{@{}l>{\raggedright\arraybackslash}Xcccc@{}}
\toprule
Block & Content (exact header) & A & B & C & D \\
\midrule

\texttt{demo}
& ``A \{age\}-year-old \{sex\} patient in intensive care.''
& \cyes & \cyes & \cyes & \cyes \\

\texttt{codes}
& ``Coded diagnoses on record: \ldots'' (prior admissions; $\leq 25$ names),
or ``No coded diagnoses on record.''
& \cyes & \cyes & \cyes & \cyes \\

\texttt{meds}
& ``Current medications (\{n\}): \ldots'' (first 48\,h; $\leq 40$ names)
& \cyes & \cyes & \cyes & \cyes \\

\texttt{labs}
& ``Laboratory values, first 24 hours: creatinine \{min--max\}, BUN \ldots''
(up to 17 analytes; legacy header, see below)
& & \cyes & & \cyes \\

\texttt{report}
& ``Radiology report for the chest film describes: \ldots''
($\leq 20$ UMLS finding names)
& & \cyes & & \cyes \\

\texttt{imaging}
& ``Chest radiograph, automated image analysis reports: \ldots''
(BiomedCLIP findings)
& & \cyes & & \cyes \\

\texttt{kg}
& Candidate packet, App.~\ref{supp:kg}
& & & \cyes & \cyes \\

\bottomrule
\end{tabularx}
\end{table}

For exact reproduction, the frozen prompts retain the legacy
``first 24 hours'' laboratory header, although all laboratory values were
computed from the prespecified first-48-hour feature window; this header
was identical across experimental conditions.

The \texttt{labs}, \texttt{report}, and \texttt{imaging} blocks are omitted when empty
(no chest film in the window: 380 items lack a report block and 1{,}458 an imaging block),
whereas an empty \texttt{codes} block is rendered as the fixed sentence above.
Because a positive and its matched negative share the admission, patient-side block
presence is identical within every pair.
The \texttt{kg} block is never empty: every negative has a non-empty packet
(the presence-leak control of App.~\ref{app:benchcons}).

\subsection{Knowledge-graph packet serialization}
\label{supp:kg}

The main experiments use the prose-list serialization.
A packet is a header sentence naming the candidate drug (its PrimeKG name) and the
relation family, followed by the complete list of related PrimeKG concept names,
one per line, sorted alphabetically:
\begin{PromptVerbatim}
Facts from a curated biomedical knowledge graph. {Drug} {relation phrase}
(complete list, {k} entries):
  {concept 1}
  {concept 2}
  ...
\end{PromptVerbatim}
with relation phrases ``is contraindicated in the following conditions''
(contra-labs, contra-report), ``has the following documented adverse effects''
(drug-effect), and ``acts on proteins implicated in the following conditions''
(mechanism).
For mechanism, the retriever applies the two-hop drug$\to$protein$\to$disease template
of \S\ref{sec:method}; the packet lists the disease endpoints of that path and does not
name the intermediate protein, so the model joins the endpoint to the patient's finding.
The typed-triple rendering compared in Table~\ref{tab:multimodel} carries the same
entries as
\begin{PromptVerbatim}
Knowledge-graph retrieval results (complete; {k} triples):
(:Drug {name: "{Drug}"})-[:CONTRAINDICATED_IN]->(:Condition {name: "{concept 1}"})
...
\end{PromptVerbatim}
with relation names \texttt{CONTRAINDICATED\_\allowbreak IN}, \texttt{HAS\_\allowbreak ADVERSE\_\allowbreak EFFECT}, and
\texttt{TARGETS\_\allowbreak PROTEIN\_\allowbreak IMPLICATED\_\allowbreak IN}, exactly as written by the v2 preparation step
into the frozen item file.

\subsection{Question wording by family}
\label{supp:questions}

The candidate drug is inserted as the MIMIC prescription string
(for example ``doxycycline hyclate'' or ``0.9\% sodium chloride'').
\begin{PromptVerbatim}
contra-labs, contra-report:
  Question: Is {drug} contraindicated for this patient, given their documented
  condition? Answer yes or no.
drug-effect:
  Question: Is at least one of this patient's documented findings a listed adverse
  effect of {drug}? Answer yes only if you can identify a specific finding that
  appears in the adverse-effect list; otherwise answer no. Answer yes or no.
mechanism:
  Question: Does {drug} act on a protein implicated in a condition this patient
  has? Answer yes or no.
\end{PromptVerbatim}

\subsection{Worked examples}
\label{supp:examples}

Each family is illustrated by one positive item and its same-admission matched
negative (same record, different candidate drug).
Only the blocks that carry the answer are shown in full; patient-level values that
are not already reported in the main text are replaced by bracketed placeholders,
and omitted list entries are counted.
The positive's answer-bearing entries are marked with $\blacktriangleleft$.
Item numbers index the frozen 3{,}200-item benchmark file.

\paragraph{Contra-labs (items 658 / 1720; Figure~\ref{fig:factorial-conditions}).}
Shared record (A--D):
\begin{PromptVerbatim}
A 83-year-old female patient in intensive care.

No coded diagnoses on record.

Current medications (33): [31 entries omitted; the list includes doxycycline
  hyclate and vancomycin].
\end{PromptVerbatim}
Patient modalities (B, D):
\begin{PromptVerbatim}
Laboratory values, first 24 hours: creatinine 4.3-4.4, BUN 91-108,
[14 further analytes omitted].                                        <|
Radiology report for the chest film describes: [15 finding names omitted].
Chest radiograph, automated image analysis reports: Pneumonia, Pulmonary Edema.
\end{PromptVerbatim}
Packet (C, D), positive candidate:
\begin{PromptVerbatim}
Facts from a curated biomedical knowledge graph. Vancomycin is contraindicated in
the following conditions (complete list, 6 entries):
  agranulocytosis
  auditory system disease
  hearing disorder
  kidney disease                                                       <|
  neutropenia
  pyoureter
\end{PromptVerbatim}
Packet (C, D), matched negative:
\begin{PromptVerbatim}
Facts from a curated biomedical knowledge graph. Doxycycline is contraindicated in
the following conditions (complete list, 3 entries):
  Clostridium difficile colitis
  dyskinesia of esophagus
  perinatal necrotizing enterocolitis
\end{PromptVerbatim}
Questions: ``Is vancomycin contraindicated \ldots'' (gold \texttt{yes}) and
``Is doxycycline hyclate contraindicated \ldots'' (gold \texttt{no}).
The creatinine value satisfies the renal-impairment rule ($>2.0$), the coded
history is empty, and only the vancomycin packet lists a renal condition;
the state is never named in the prompt.
Logged scores (Qwen3.5-35B): positive A 0.011, B 0.026, C 0.002, D 0.651;
negative A 0.033, B 0.039, C 0.005, D 0.025.

\paragraph{Contra-report (items 3101 / 2975).}
\begin{PromptVerbatim}
A [age]-year-old [sex] patient in intensive care.

No coded diagnoses on record.

Current medications (40): [40 entries omitted; propofol is listed, omeprazole falls
  beyond the 40-name cap and is named only in the question].

Laboratory values, first 24 hours: [15 analytes omitted].
Radiology report for the chest film describes: [3 findings omitted],
  Congestive heart failure, [2 findings omitted].                       <|
Chest radiograph, automated image analysis reports: Multiple pulmonary nodules.
\end{PromptVerbatim}
Positive packet (omeprazole, 8 entries): bone fracture, chronic atrophic gastritis,
congestive heart failure $\blacktriangleleft$, gallbladder disease, liver disease,
liver failure, osteoporosis, postmenopausal osteoporosis.
Negative packet (propofol, 6 entries): epilepsy, familial hyperlipidemia,
hypotensive disorder, inherited lipid metabolism disorder, intracranial
arteriosclerosis, respiratory failure.
The report finding aligns to PrimeKG \emph{congestive heart failure}, which appears
only in the omeprazole packet; no entry of either packet is a near-synonym of any
other visible finding.

\paragraph{Drug-effect (items 1662 / 1285).}
\begin{PromptVerbatim}
A [age]-year-old [sex] patient in intensive care.

No coded diagnoses on record.

Current medications (33): [31 entries omitted; the list includes 0.9% sodium
  chloride and chlorhexidine gluconate 0.12% oral rinse].

Laboratory values, first 24 hours: [15 analytes omitted].
Radiology report for the chest film describes: Hypervolemia (finding),
  Left pleural effusion.                                               <|
\end{PromptVerbatim}
(No imaging block: the study had no finding above threshold, so the block is omitted.)
Positive packet: ``Sodium chloride has the following documented adverse effects
(complete list, 22 entries)'': Agitation, Coma, Confusion, Congestive heart failure,
Dehydration, Drooling, Edema, Epiphora, Excessive daytime somnolence, Excessive
salivation, Fever, Headache, Hypernatremia, Hypervolemia $\blacktriangleleft$,
Loss of consciousness, Pain, Respiratory failure, Seizure, Tachycardia,
Thrombophlebitis, Venous thrombosis, Vertigo.
Negative packet: ``Chlorhexidine has the following documented adverse effects
(complete list, 25 entries)'': Abnormal joint morphology, Anaphylactic shock,
Arthralgia, Arthropathy, Back pain, Bronchitis, Cough, Dysmenorrhea, Erythema,
Gingival bleeding, Gingival overgrowth, Glossitis, Headache, Myalgia, Oral ulcer,
Pain, Paresthesia, Recurrent aphthous stomatitis, Recurrent fractures, Rhinitis,
Scaling skin, Sensory impairment, Sinusitis, Stomatitis, Xerostomia.
This pair shows the length matching: the negative list is longer than the
positive one (rendered-character ratio 1.10, within the 0.85--1.18 tolerance), so list
length cannot stand in for the answer.

\paragraph{Mechanism (items 101 / 2202).}
\begin{PromptVerbatim}
A [age]-year-old [sex] patient in intensive care.

No coded diagnoses on record.

Current medications (24): [22 entries omitted; the list includes alteplase and
  chlorhexidine gluconate 0.12% oral rinse].

Laboratory values, first 24 hours: [15 analytes omitted].
Radiology report for the chest film describes: [5 findings omitted],
  Pulmonary Embolism, [2 findings omitted].                            <|
\end{PromptVerbatim}
Positive packet: ``Alteplase acts on proteins implicated in the following conditions
(complete list, 62 entries)'', which include \emph{pulmonary embolism (disease)}
$\blacktriangleleft$ (reached through alteplase's target fibrinogen alpha chain, FGA, which is
not shown in the prompt); the other 61 entries are omitted here.
Negative packet: ``Chlorhexidine acts on proteins implicated in the following
conditions (complete list, 71 entries)''; none of the 71 entries aligns to, or is a
near-synonym of, a finding visible in the record.
The protein hop is composed by the retriever.
Typed one-hop retrieval (R3) returns drug$\to$protein edges but never the
protein$\to$disease hop, so its mechanism path recall is 0 at every budget in the
frozen test analysis (App.~\ref{app:mimic-retrieval} reports the pooled recall).

\subsection{Five-candidate ranking prompts}
\label{supp:ranking}

The ranking benchmark of \S\ref{sec:external} reuses the patient-side blocks of
App. \ref{supp:blocks} unchanged (same renderer and caps, including the 40-name medication
cap; test items list 13--40 medications, median 37).
For the per-candidate scoring, the record is followed by a candidate line and one
question per candidate:
\begin{PromptVerbatim}
Candidate drugs received by this patient: 1. {drug 1}; 2. {drug 2}; ...; 5. {drug 5}.

Question: Does this patient have a documented condition in which {drug i} is
contraindicated, or a documented finding that is a listed adverse effect of {drug i}?
Answer yes or no.
\end{PromptVerbatim}
Condition D in the R4q formulation includes only the queried candidate's
contraindication and adverse-effect packets (both relation families, same prose
serialization as App. \ref{supp:kg}); the earlier five-packet rendering included all
five candidates' packets in every question.
The five candidates are ranked by their yes-probabilities.

\end{document}